%% file: TNNLS-2020-S-14911.tex
\documentclass[journal]{IEEEtran}

\usepackage{graphicx}
\usepackage{xcolor}
\usepackage{subfigure}
\usepackage{booktabs} 

\usepackage{hyperref}
\usepackage[noadjust]{cite}
\usepackage{amsmath,amssymb,amsthm}

\begin{document}

\title{On Information Plane Analyses of Neural Network Classifiers -- A Review}

\author{Bernhard C. Geiger,~\IEEEmembership{Senior Member,~IEEE}%
\thanks{The work of Bernhard C. Geiger was supported by the iDev40 project. The iDev40 project has received funding from the ECSEL Joint Undertaking (JU) under grant agreement No 783163. The JU receives support from the European Union’s Horizon 2020 research and innovation programme. It is co-funded by the consortium members, grants from Austria, Germany, Belgium, Italy, Spain and Romania.
The Know-Center is funded within the Austrian COMET Program - Competence Centers for Excellent Technologies - under the auspices of the Austrian Federal Ministry for Climate Action, Environment, Energy, Mobility, Innovation and Technology, the Austrian Federal Ministry of Digital and Economic Affairs, and by the State of Styria. COMET is managed by the Austrian Research Promotion Agency FFG.}%
\thanks{Bernhard C. Geiger is with Know-Center GmbH, Graz, Austria. (\mbox{e-mail:} \mbox{geiger@ieee.org})}}

\newcommand{\dom}[1]{\mathcal{#1}}
\newcommand{\Prob}[1]{\mathbb{P}(#1)}
\newcommand{\dataset}{\mathcal{D}}
\newcommand{\mutinf}[1]{I(#1)}
\newcommand{\mutest}[1]{\hat{I}(#1)}
\newcommand{\ent}[1]{H(#1)}
\newcommand{\entest}[1]{\hat{H}(#1)}
\newcommand{\diffent}[1]{h(#1)}

\maketitle

\begin{abstract}
 We review the current literature concerned with information plane analyses of neural network classifiers. While the underlying information bottleneck theory and the claim that information-theoretic compression is causally linked to generalization are plausible, empirical evidence was found to be both supporting and conflicting. We review this evidence together with a detailed analysis of how the respective information quantities were estimated. Our survey suggests that compression visualized in information planes is not necessarily information-theoretic, but is rather often compatible with geometric compression of the latent representations. This insight gives the information plane a renewed justification.

Aside from this, we shed light on the problem of estimating mutual information in deterministic neural networks and its consequences. Specifically, we argue that even in feed-forward neural networks the data processing inequality need not hold for estimates of mutual information. Similarly, while a fitting phase, in which the mutual information between the latent representation and the target increases, is necessary (but not sufficient) for good classification performance, depending on the specifics of mutual information estimation such a fitting phase need not be visible in the information plane.
\end{abstract}
\begin{IEEEkeywords}
 information bottleneck, information plane analysis, deep learning, information theory, deep neural networks
\end{IEEEkeywords}

\section{Introduction and Motivation}
\label{sec:intro}
The information bottleneck (IB) theory of deep learning, initially proposed in~\cite{Tishby_DLIB_ITW}, suggests that a learned latent representation in a neural network (NN) should contain all information from the input required for estimating the target -- but not more than this required information. The NN should be \emph{fit} to the target and simultaneously \emph{compress} all irrelevant information to prevent overfitting. If we identify input, target, and latent representations with random variables (RVs) $X$, $Y$, and $L$ respectively, then $L$ should be a \emph{minimal sufficient statistic} for $Y$ obtained from $X$. In information-theoretic terms, this is equivalent to finding a latent representation $L$ that minimizes the mutual information $\mutinf{X;L}$ with the input $X$ while satisfying $\mutinf{Y;L}=\mutinf{X;Y}$. Since the minimizing $L$ may not be obtainable with a NN of a given architecture, one can relax the problem to the IB problem:
\begin{equation}\label{eq:IB}
 \min_{P_{L|X}} \mutinf{X;L} -\beta\mutinf{Y;L}
\end{equation}
where $\beta$ trades between preserving target-relevant information and compressing target-irrelevant information and where the feasible set depends on the NN architecture.

\begin{figure}[t]
\begin{center}
\centerline{\includegraphics[width=\columnwidth,trim={0cm 0.25cm 0cm .20cm}, clip]{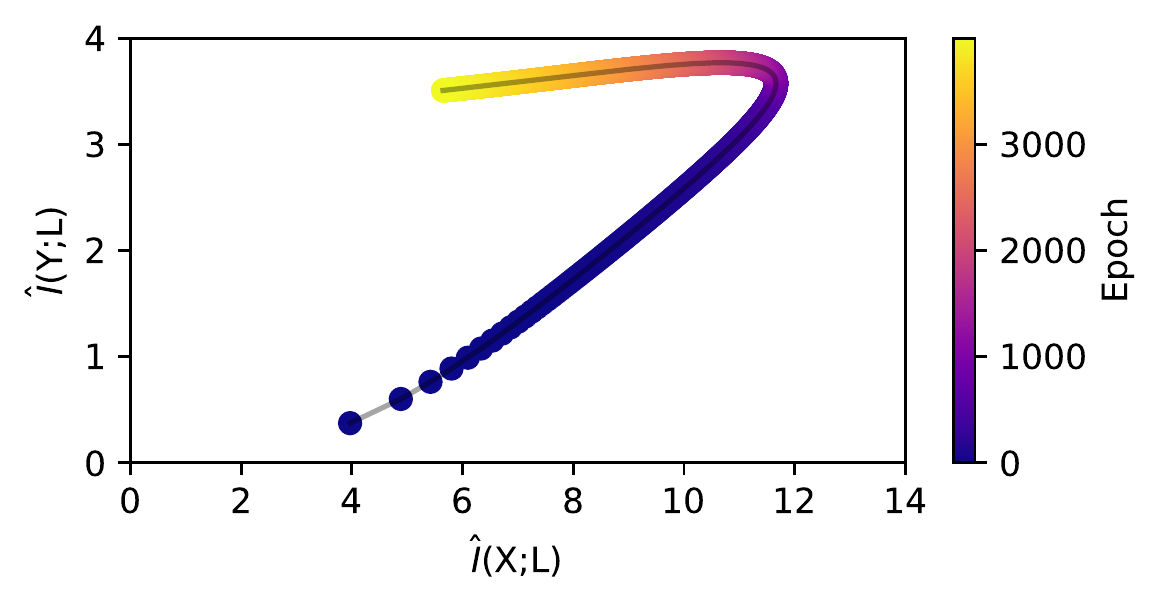}}
\caption{Faux information plane for one NN layer. Two phases are visible: a fitting phase during which the mutual information estimate $\mutest{Y;L}$ increases, and a subsequent phase of compression (decrease of $\mutest{X;L}$) and overfitting (small decrease of $\mutest{Y;L}$). See text for details.}
\label{fig:faux}
\end{center}
\vskip -0.2in
\end{figure}

\input{table.tex}

Based on the IB theory, the authors of~\cite{Tishby_BlackBox} popularized the analysis of the \emph{information plane} (IP), in which estimates of the two quantities $\mutinf{X;L}$ and $\mutinf{Y;L}$ are the coordinate axes (see Fig.~\ref{fig:faux}). The IP is used to visualize how the estimates of $\mutinf{X;L_t}$ and $\mutinf{Y;L_t}$ change with the training epoch $t$; e.g., \emph{fitting} to the target is indicated by an increase of the estimate of $\mutinf{Y;L_t}$ as $t$ increases, and \emph{compression} is characterized by a decrease of the estimate of $\mutinf{X;L_t}$. For example, the authors of~\cite{Tishby_BlackBox} observed that training a NN with stochastic gradient descent (SGD) is characterized by a short fitting phase followed by a long compression phase, which the authors claimed to be connected to generalization. This enticing idea, that the reason behind good generalization performance can simply be read off a chart, led to the belief that IP analyses could reveal more about the inner workings of NNs -- that they may ``open the black box of deep learning''. Indeed, for the case of an input $X$ with finite alphabet, the authors of~\cite{Vera_IBRepresentation} have shown that an estimate of $\mutinf{X;L}$ from the dataset appears in a bound on the generalization error.

Several authors have subsequently performed IP analyses, applying different estimation mechanisms to different NN architectures, including unsupervised structures such as autoencoders. The presented evidence is conflicting: For example, the authors of~\cite{Saxe_IBTheory} did not observe compression at all for NNs with ReLU activation functions, and the authors of~\cite{Chelombiev_AdaptiveEstimatorsIB} observed compression sometimes at an earlier phase, sometimes at a later phase of training, depending on the initialization of the NN parameters. Also, the link between compression and generalization has been questioned,~cf.~\cite{Saxe_IBTheory}. 

It is the goal of this paper to summarize and consolidate these partly conflicting observations. To this end, we structure the paper into two parts. In the first part, we start by discussing the complications of estimating mutual information in deterministic NNs in Section~\ref{sec:estimation} and argue that IP analyses obtained with different estimators are not directly comparable. We then shed some light on the validity of the data processing inequality (DPI) in NNs and discuss potential connections between $\mutest{Y;L}$ and generalization performance in Sections~\ref{sec:DPI} and~\ref{sec:fano}, respectively. Building upon these theoretical insights, in the second part of this paper we survey the literature on IP analyses of NNs (Section~\ref{sec:analyses}). Our main observations from this survey are:
\begin{itemize}
 \item Compression in the IP is often compatible with geometric compression, i.e., with $L$ being ``small'' or densely clustered in latent space, while information-theoretic compression in the sense of reducing $\mutinf{X;L}$ can often be ruled out due to the complications mentioned in Section~\ref{sec:estimation}.
 \item The convolutional layers of convolutional NNs (CNNs) appear to behave as invertible functions at all training epochs, at least on the dataset.
 \item Regardless of whether the IP displays information-theoretic or geometric compression, evidence presented for the \emph{presence} of a causal link between compression in the IP and good generalization performance is less convincing than the evidence presented for its \emph{absence}.
\end{itemize}

Based on these observations and the preceding theoretical considerations, we conclude that the requirement that $L$ shall be a \emph{minimal} sufficient statistic appears unnecessary. This in turn requires us to reassess the validity and value of IP analyses, which we do in Section~\ref{sec:summary}. More specifically, we argue that the IP can shed light on the geometric effects occurring during training a NN, given that the specifics of mutual information estimation are properly taken into account.

\textbf{Related Work:} Some of the conclusions we draw in this work have also appeared elsewhere: For example, the discrepancy between mutual information in deterministic NNs and its estimates was investigated in~\cite{Amjad_LearningRepresentations,Goldfeld_Estimating,Abrol_MFI}. More generally, the authors of~\cite{Yu_MatrixEntropies,Yu_AE} mentioned that estimates of mutual information need not necessarily inherit all properties from the probabilistic definition of mutual information. Also the claim that compression in the IP can be explained by geometric effects is not new, cf.~\cite{Schiemer_RevisitingIP}, which investigated binning estimators for classification scenarios, and~\cite[Fig.~8]{Yu_AE}, which shows geometric compression in the latent space of a stacked autoencoder. Another excellent paper on the interplay between geometric compression and the IP is~\cite{Goldfeld_Estimating}, which argues that information-theoretic compression in noisy NNs is linked to clustering, that a binning estimator of the entropy of a latent representation is an adequate estimator of geometric clustering, and that CNNs tend to compress geometrically, but that this is not visible in the IP due to the high dimensionality of the latent representations. Rather than arguing that clustering is the phenomenon of interest in training NNs~\cite[Sec.~6]{Goldfeld_Estimating}, we claim that geometric compression (shrinking \emph{or} clustering) is a phenomenon that is consistent with compression observed in the IP for a wide range of mutual information estimators.

Finally, it is worth mentioning two recent surveys on the IB principle applied to deep learning: First,~\cite{Hafez-Kolahi_Review} reviewed the main findings of~\cite{Tishby_BlackBox,Saxe_IBTheory} as well as the connection between the IB principle, information dropout~\cite{Achille_InfoDropout}, and variational autoencoders~\cite{Kingma_VAE}. And second,~\cite{Goldfeld_Review} reviewed the application of the IB functional as a cost function for NN training~\cite{Alemi_DVIB,Achille_InfoDropout}, summarized the main findings of~\cite{Tishby_BlackBox,Saxe_IBTheory,Goldfeld_Estimating}, and presented the operational meaning of the IB functional in communications and information theory. In contrast to these, the focus of our survey is exclusively on IP analyses, aiming to provide a comprehensive picture and consolidating the currently available results.

IP analyses were proposed to deepen our understanding of how a NN trains and what (geometric) properties of a latent representations make it yield high classification performance. As such, IP analyses and the IB theory of NNs can help explaining NNs to the machine learning expert. They thus fall into the same category of theories and tools as the lottery ticket hypothesis~\cite{Frankle_Lottery}, analyses regarding the importance of individual neurons~\cite{Deepmind_Cats}, or investigations of NNs based on the neural tangent kernel~\cite{Jacot_NTK}. A category separate from this contains tools and theories aiming to explain NNs and its decisions to the end user. This category is typically connected with the notion of interpretability, explainable machine learning, or explainable artificial intelligence. Tools from this category often rely on visualizations, extracted rule sets, or other human-graspable concepts to explain individual decisions of the NNs. For surveys for this latter category, we refer the interested reader to~\cite{Samek_xAI,Montavon_Interpretability,Zhang_Interpretability,Li_Interpretability,Adadi_Interpretability}.

\textbf{Considered Literature and Scope:} We consider work investigating the behavior of estimates of $\mutinf{X;L_t}$ and $\mutinf{Y;L_t}$ during training of NNs; see Table~\ref{tab:literature} for an overview. We do not consider information-theoretic analyses of trained NNs, such as~\cite{Davis_NIF,Choi_StatsIT,Amjad_NeuronImportance}; works concerned with the mutual information between the network input (data) $X$ ($\dataset$) and the network weights, such as~\cite{Achille_Emergence}; IP analyses of NNs trained for regression or of autoencoders, such as~\cite{Yu_AE,Tapia_AE,Lee_AEs}; and the interesting body of literature concerned with information-theoretic objectives for NN training, such as~\cite{Alemi_DVIB,Achille_InfoDropout,Kolchinsky_NLIB}. However, the fast pace at which this scientific field progresses requires us to consider unpublished work in addition to work that has passed peer-review. That such an approach is accepted by the scientific community sees evidence in the fact that~\cite{Tishby_BlackBox} has accumulated more than 700 citations, but at the time of writing has not yet passed peer-review.

\textbf{Notation:} We let $X$ and $Y$ denote RVs representing the features and target of a classification problem. Typically, $X$ has a continuous distribution on a high-dimensional space, while $Y$ has a discrete distribution on a finite set of classes. Every dataset $\dataset=\{(x_1,y_1),\dots,(x_N,y_N)\}$ is assumed to contain independent realizations of the joint distribution $P_{X,Y}$ of $X$ and $Y$. The features $X$ are the input to a NN. We describe each NN by the widths of the hidden layers and let the input and output dimensions be defined from the context. For example, $1024-512-256$ is a multi-layer perceptron (MLP) with three hidden layers of widths $1024$, $512$, and $256$, respectively. The output of a layer or filter inside the NN at epoch $t$ defines the latent representation $L_t$; we suppress the epoch index $t$ for the sake of readability. We use indices, e.g., $L_i$, $L_j$,\dots, to refer to latent representations of different layers. We call a NN \emph{deterministic} if it implements a function $f$ from the input $X$ to the latent representation $L$, i.e., if $L=f(X)$. We call a NN \emph{stochastic} if it implements a conditional distribution. We let $\diffent{\cdot}$, $\mutinf{\cdot;\cdot}$, $\ent{\cdot}$, $\mutest{\cdot;\cdot}$, and $\entest{\cdot}$ denote differential entropy, mutual information, entropy, and their estimates, respectively.

\section{Estimating Mutual Information in Neural Networks}
\label{sec:estimation}
Since the IP aims to display how $\mutinf{X;L}$ and $\mutinf{Y;L}$ change during training, these quantities need to be estimated from a dataset $\dataset$. This is at least theoretically possible if the quantities $\mutinf{X;L}$ and $\mutinf{Y;L}$ are finite. For example,  $\mutinf{Y;L}$ is finite for classification tasks in which $Y$ is a RV on a finite set $\mathcal{Y}$ of classes. Thus, one can reasonably assume that $\mutest{Y;L}\approx\mutinf{Y;L}$ if the estimator is adequately parameterized. For example, if $Q(\cdot)$ is a quantizer, then the plug-in estimate for $\mutinf{Y;L}$ obtained from dataset $\dataset$ yields
\begin{subequations}\label{eq:plugin}
\begin{multline}
 \mutest{Y;L}
  = \sum_{q\in\mathsf{range}(Q)}\sum_{y\in\dom{Y}} \hat{p}_{Y,Q(L)}(y,q) \log \frac{\hat{p}_{Y,Q(L)}(y,q)}{\hat{p}_{Q(L)}(q)\hat{p}_{Y}(y)}\label{eq:plugin:Y}
\end{multline}
where $\hat{p}_{Y,Q(L)}(y,q)=|\{i{:}\ Q(f(x_i))=q, y_i=y\}|/|\dataset|$, and where $\hat{p}_{Q(L)}$ and $\hat{p}_{Y}$ are obtained by marginalizing $\hat{p}_{Y,Q(L)}$. Then, $\mutest{Y;L}\approx\mutinf{Y;L}$ if $Q$ has appropriate bin size and if $\dom{D}$ is sufficiently large.

There are settings in which also $\mutinf{X;L}$ is finite, cf.~\cite[Sec.~5]{Amjad_LearningRepresentations}. For example, $\mutinf{X;L}$ is finite if the NN under consideration has only finitely many activation values (such as binary or ternary NNs) or if the latent representation $L=\hat{Y}$ is the (finite-alphabet) class-estimate of the NN. Also, if the NN is stochastic, as in scenarios where $f$ is a probability distribution parameterized by a NN, $\mutinf{X;L}$ may be finite. In these cases, an IP analysis is unproblematic if the estimators can be parameterized such that their estimates are close to the true (finite) quantities. 

\begin{figure*}
 \subfigure[$\mutest{X;L}\approx 4$, $\mutest{Y;L}=2$]{\includegraphics[width=0.25\textwidth]{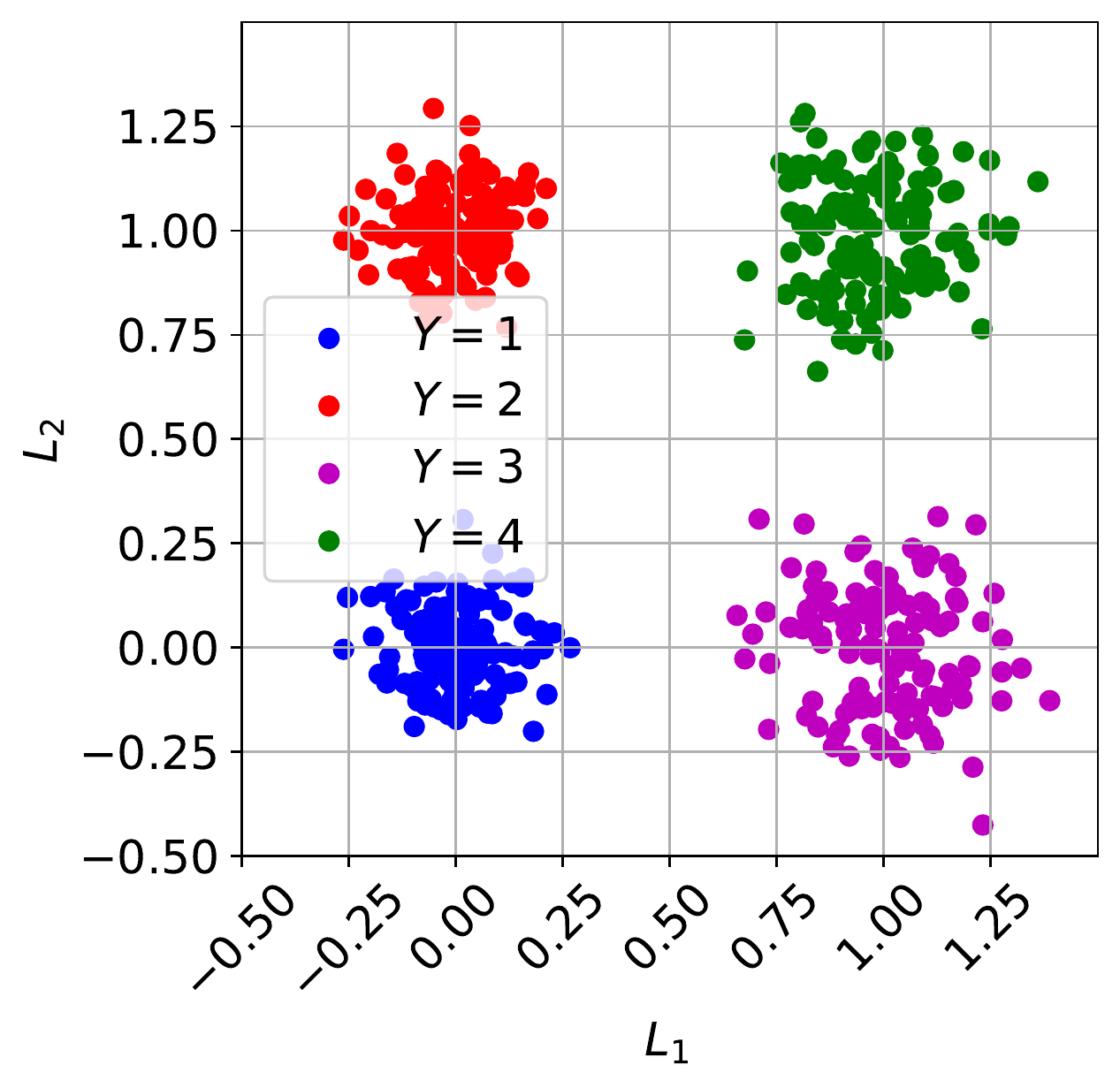}}\hfill
 \subfigure[$\mutest{X;L}= 2$, $\mutest{Y;L}=2$]{\includegraphics[width=0.25\textwidth]{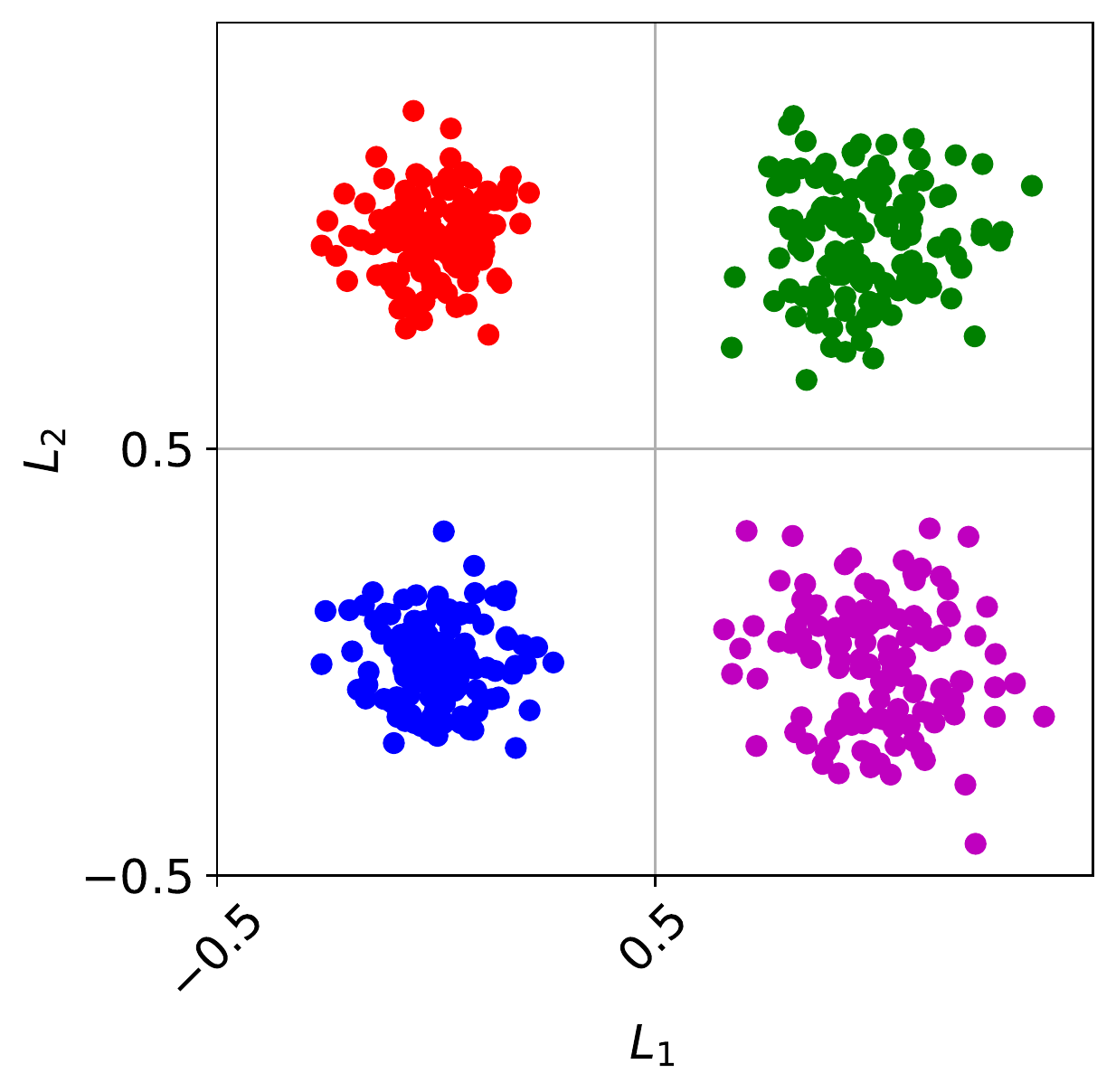}}\hfill
 \subfigure[$\mutest{X;L}\approx 0$, $\mutest{Y;L}\approx 0$]{\includegraphics[width=0.25\textwidth]{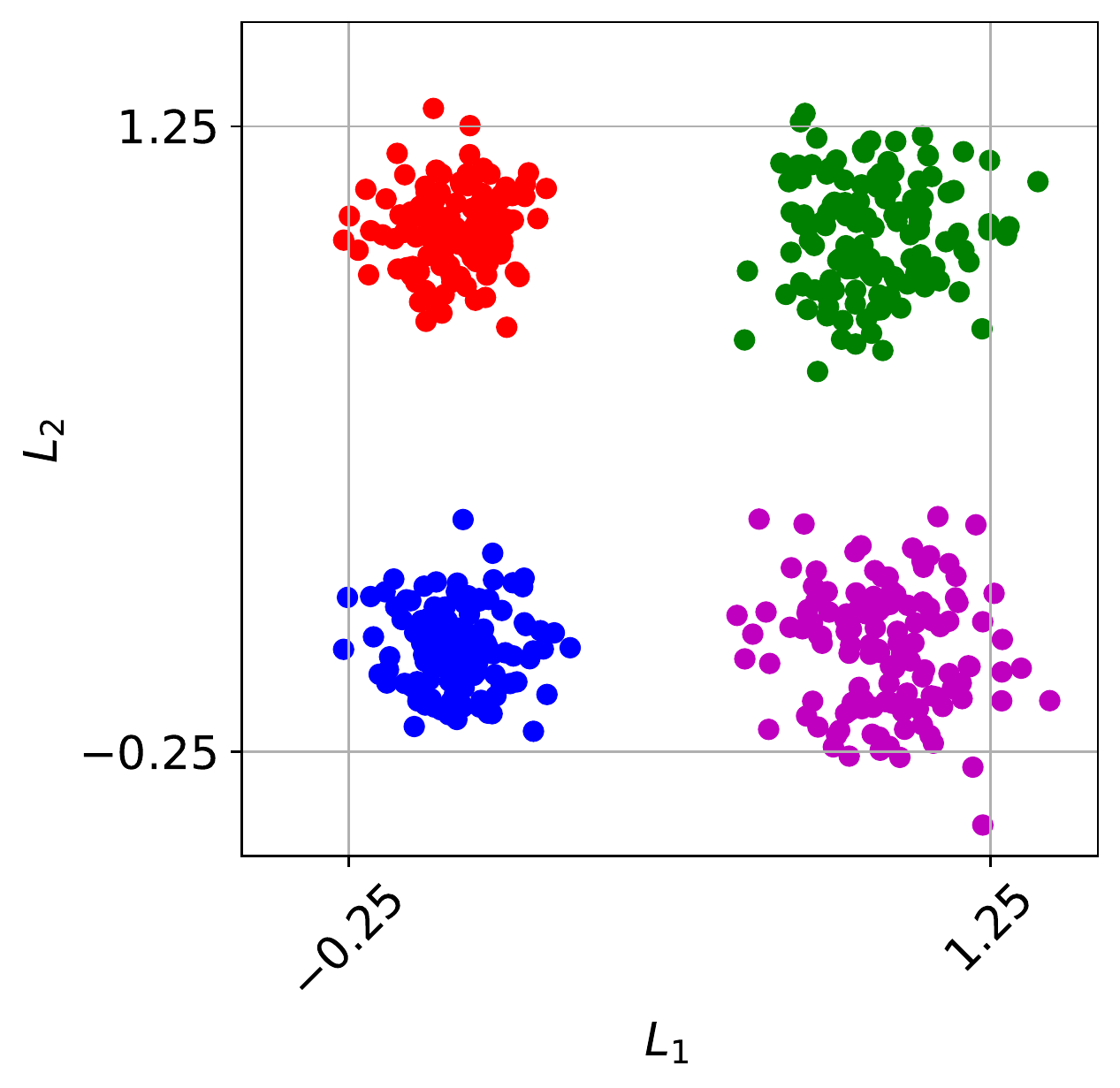}}\hfill
 \subfigure[$\mutest{X;L}\approx \log|\dataset|$, $\mutest{Y;L}=2$]{\includegraphics[width=0.25\textwidth]{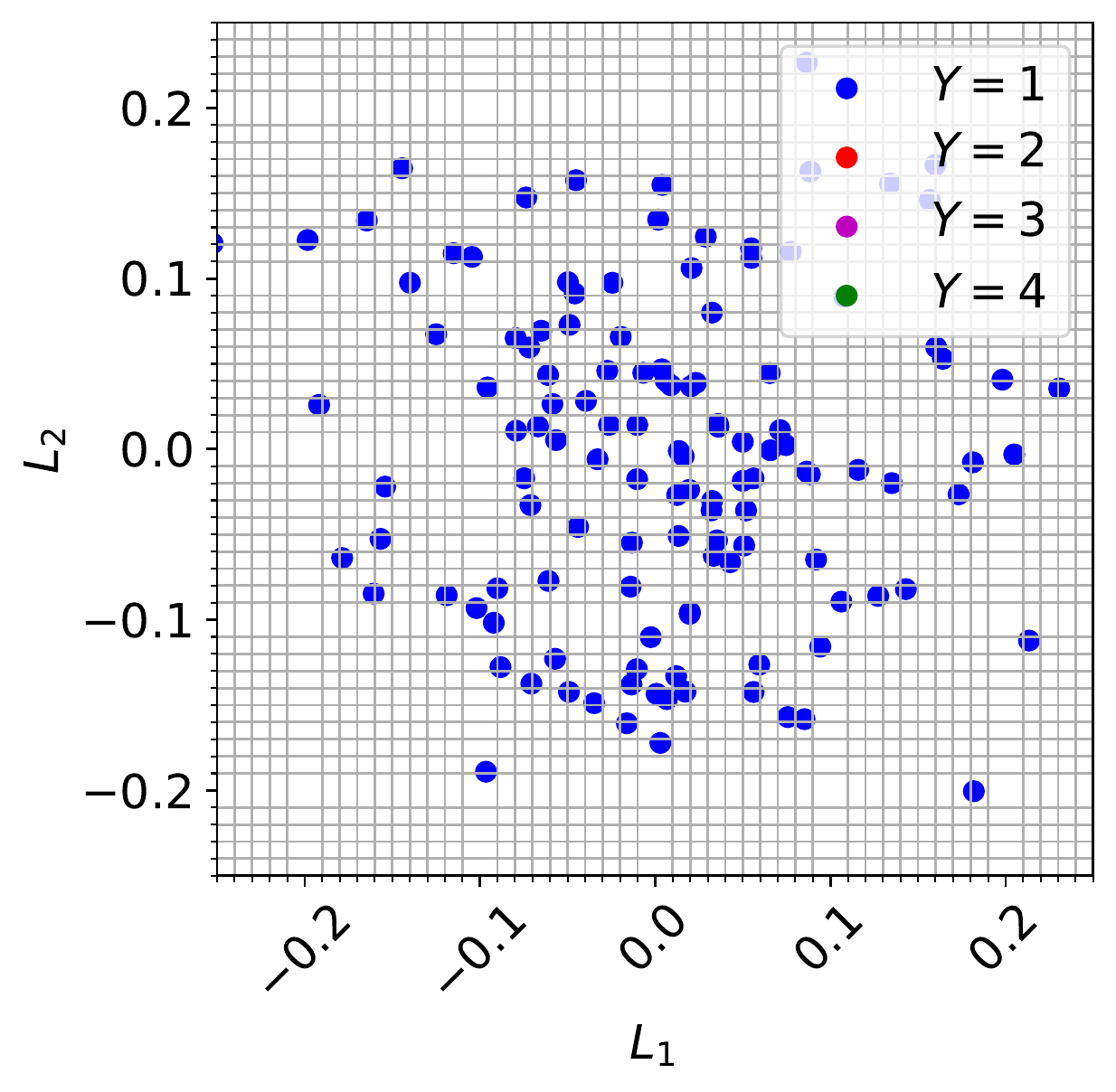}}
 \caption{Geometric influence on binning estimates of mutual information. A dataset with $|\dataset|=512$ and four equiprobable classes is considered. The images depict a faux two-dimensional latent representation $L$ for a deterministic NN at some epoch during training, grid lines indicate the bins of the quantizer $Q$. Mutual information estimates $\mutest{X;L}=\entest{Q(L)}$ and $\mutest{Y;L}$ are obtained via~\eqref{eq:plugin}. (a,b) If the bin size is larger than the spacing between data points of the same class, but smaller than the spacing between clusters, then $\mutest{X;L}$ is small compared to the dataset size and $\mutest{Y;L}$ approximates $\mutinf{X;Y}$, which we assume to be $\ent{Y}=2$ in this case. (c) If the bin size is larger than the spacing between clusters, then most of the data points fall into one bin and both estimates for mutual information are close to zero. (d) If the bin size is too small, then most of the data points fall into a separate bin, yielding $\mutest{X;L}\approx|\dataset|$ and $\mutest{Y;L}\approx\ent{Y}$. Note that a fixed increase of the bin size has an equivalent effect as a fixed scaling of $L$ with a number smaller than $1$.}
 \label{fig:binning}
\end{figure*}

However, if the NN is deterministic, estimating the mutual information $\mutinf{X;L}$ (which in this case coincides with the entropy $\ent{L}$) is problematic, if not futile. To make this clear, we discuss two common assumptions regarding the distribution $P_{X}$ of the features $X$. 

In the first case, we assume that the distribution $P_X$ is continuous.\footnote{A discrete distribution on a high-dimensional finite space is appropriately modeled as being continuous. E.g., for image classification, the pixel $X_i$ may have a discrete distribution on $\{0,\dots,255\}^3$ that is modeled as a continuous distribution on $[0,1]^3$.} If the NN is such that also $L$ has a continuous distribution, then one can easily show that $\mutinf{X;L}=\infty$~\cite{Saxe_IBTheory,Goldfeld_Estimating}. More rigorously,~\cite[Th.~1]{Amjad_LearningRepresentations} shows that $\mutinf{X;L}=\infty$ for continuous $X$ and many common activation functions (incl. tanh, sigmoid, or leaky ReLU), even if $L$ is not continuous. 

In the second case, we assume that the features follow the empirical distribution of the dataset $\dataset$, i.e., the distribution $P_X$ has a point mass at the position of every sample from the dataset $\dataset$. In this case, most NNs will map the dataset bijectively, i.e., if the features of two samples in the dataset are distinct, then so will be all their latent representations\footnote{More precisely, this holds for almost all weight matrices; i.e., it may not hold for weight matrices drawn from a set that has zero Lebesgue measure.} -- this is a simple consequence of the fact that $|\dataset|$ is usually much smaller than the cardinality of the feature space and the fact that most activation functions have a strictly monotonic part. Under this assumption on $P_X$, one has $\mutinf{X;L}=\log|\dataset|$ and $\mutinf{Y;L}=\mutinf{Y;X}$, cf.~\cite{Goldfeld_Estimating} and~\cite[Sec.~4.2]{Amjad_LearningRepresentations}.

In contrast, an \emph{estimator} of mutual information typically yields different results: For example, if $Q(\cdot)$ is a quantizer with large bin size and if $L$ is low-dimensional, the plug-in estimate for $\mutinf{X;L}=\ent{L}$ obtained from dataset $\dataset$ yields
\begin{multline}
 \mutest{X;L}
 = -\sum_{q\in\mathsf{range}(Q)} \hat{p}_{Q(L)}(q) \log \hat{p}_{Q(L)}(q)\ll \log|\dataset|\label{eq:plugin:X}
\end{multline}
\end{subequations}
where $\hat{p}_{Q(L)}(q)=|\{i{:}\ Q(f(x_i))=q\}|/|\dataset|$. We illustrate this and other possible outcomes of such a binning scheme in Fig.~\ref{fig:binning}. In the extreme case where $L$ and $Q$ are such that all data points fall into a single quantizer bin, this estimate for mutual information is zero. In the case where the NN clusters data points according to their class membership and if these clusters all fall into different quantizer bins, this estimate will yield $\mutest{X;L}=\entest{Y}$. Finally, if the bin size is small or if $L$ is high-dimensional, as in a CNN, then one may observe $\mutest{X;L}\approx\log|\dataset|$ because almost all data points fall into different bins; cf.~\cite[Fig.~15]{Saxe_IBTheory} and~\cite[Fig.~1]{Goldfeld_Estimating}. From this becomes apparent that the definition of the quantizer $Q$ has profound effect on the results obtained by such information-theoretic analyses, and that binning estimators are inherently influenced by geometric effects such as shrinking and clustering.

\begin{figure*}
 \subfigure[$\mutest{X;L}\approx \log 5$, $\mutest{Y;L}\approx\ent{Y}$]{\includegraphics[width=0.3\textwidth]{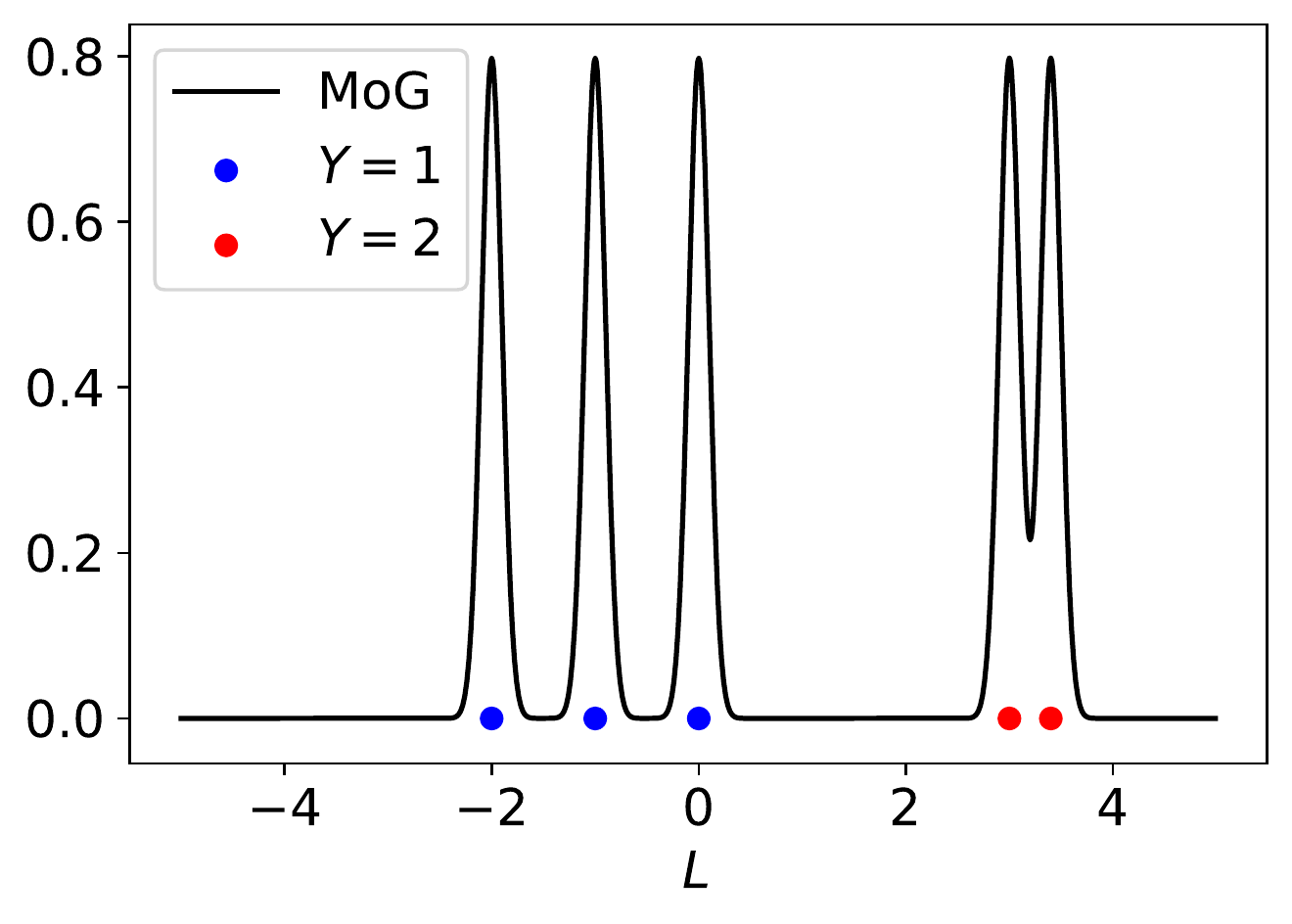}}\hfill
 \subfigure[$\mutest{X;L}\approx\ent{Y}$, $\mutest{Y;L}\approx\ent{Y}$]{\includegraphics[width=0.3\textwidth]{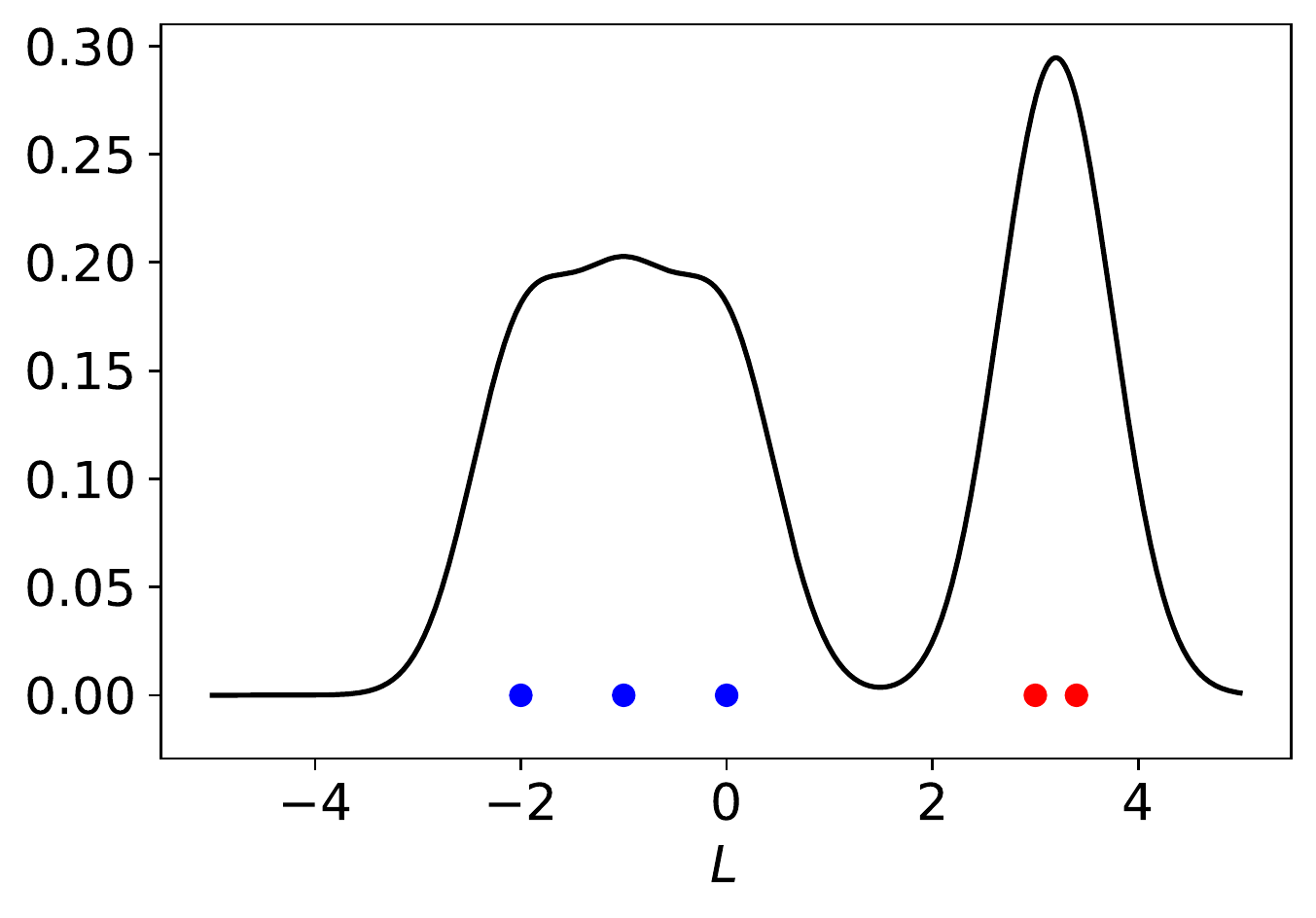}}\hfill
 \subfigure[$\mutest{X;L}\approx 0$, $\mutest{Y;L}\approx 0$]{\includegraphics[width=0.3\textwidth]{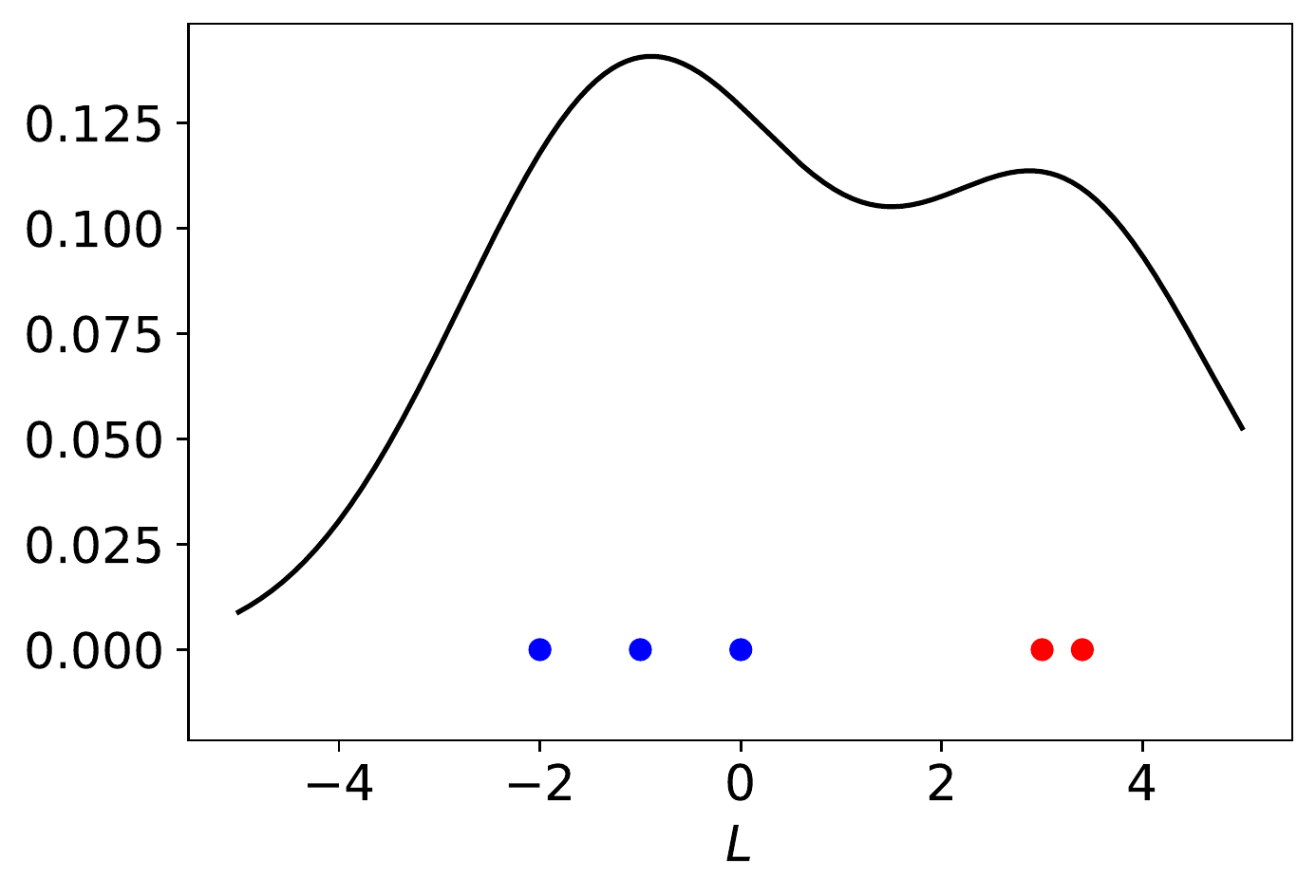}}
 \caption{Geometric influence on kernel density estimates of mutual information, or of estimates obtained via adding independent Gaussian noise $\varepsilon$ to the latent representation. In other words, $\mutest{X;L}$ and $\mutest{Y;L}$ are estimates of $\mutinf{X;L+\varepsilon}$ and $\mutinf{Y;L+\varepsilon}$, respectively. The images depict a faux one-dimensional latent representation $L$ at some epoch during training, the black lines indicate the density under a Gaussian kernel, i.e., a mixture of Gaussians. (a) If the kernel width or the variance of the Gaussian noise is small, then the mixture components are separable and $\mutest{X;L}\approx\log|\dataset|$ and $\mutest{Y;L}\approx\mutinf{X;Y}$, which we assume to be $\ent{Y}$. (b) If the kernel width is such that data points from different classes are separable, then $\mutest{X;L}$ decreases, while $\mutest{Y;L}\approx\ent{Y}$ still holds. (c) If the kernel width is such that the overlap between components for different classes is significant, then both $\mutest{X;L}$ and $\mutest{Y;L}$ tend to be small. Note that a fixed increase of the kernel width has an equivalent effect as a fixed scaling of $L$ with a number smaller than $1$.}
 \label{fig:KDE}
\end{figure*}

Similar considerations hold if $\mutest{X;L}$ is not computed using binning but, e.g., using kernel-density estimators (KDE) or via adding noise, i.e., via $\mutest{X;L}=\mutinf{X;L+\varepsilon}$, where $\varepsilon$ is an independent noise vector. Indeed, if $\varepsilon$ is Gaussian with constant covariance structure and the NN is deterministic, then $\diffent{L+\varepsilon|X}=\diffent{\varepsilon}$ and $\mutest{X;L}$ depends on the NN function $f$ exclusively via the differential entropy $\diffent{L+\varepsilon}$. If $P_{L}$ is obtained by mapping the empirical distribution of the dataset through the NN, then $L+\varepsilon$ is essentially distributed as a mixture of Gaussians (MoG, see Fig.~\ref{fig:KDE}). In the extreme case where all data points are mapped to the same point in latent space, this MoG has one component; in the other extreme where all data points are mapped far apart (w.r.t.\ the covariance of $\varepsilon$), then this MoG has $|\dataset|$ components; finally, if the data points cluster in latent space, then this MoG resembles a MoG with as many components as there are clusters, cf.~\cite{Kolchinsky_KDE},~\cite[Appendix~A2]{Kirsch_EntropyEstimator}. This shows that also KDE or noisy mutual information effectively convey a geometric picture. Since furthermore the distance between data points in latent space is measured relative to the standard deviation of $\varepsilon$, this standard deviation (or the kernel width in case of KDE) plays a similar role as the bin size for binning estimators.

The considerations in this section have an immediate consequence for IP analyses: First, since $\mutest{X;L}$ and $\mutest{Y;L}$ depend on the choice of the estimator, IPs can only be interpreted when taking the details of estimation into account. For example, results obtained via binning and KDE may be interpreted with a geometric picture in mind (e.g., clustering, scaling, etc.). Second, even if $\mutinf{\cdot;\cdot}$ is finite, we have $\mutest{\cdot;\cdot}\approx\mutinf{\cdot;\cdot}$ only if $\dataset$ is sufficiently large to allow accurate estimation, where the appropriate size of $\dataset$ depends on the choice and parameterization of the estimator. Finally, and most importantly, IPs obtained by different estimators are not directly comparable. Conflicting claims made on the basis of IP analyses can only be conflicting if the same estimators are used. If different estimators are used, conflict or agreement is only superficial, since the IPs show different things despite identical axis labelling.

\section{The Data Processing Inequality under Mutual Information Estimators}\label{sec:DPI}

For a feed-forward NN with latent representations $L_1$ through $L_m$ defined by the corresponding layers, the following Markov relation holds:
\begin{subequations}\label{eq:DPI}
\begin{equation}\label{eq:DPI:DPI}
 Y-X-L_1-L_2-\cdots-L_m
\end{equation}
 As an immediate consequence, the DPI~\cite[Th.~2.8.1]{Cover_Information} dictates that there is an ordering between several mutual information terms depicted in the IP, such as
 \begin{align}
  \mutinf{X;L_1}\ge \mutinf{X;L_2} \ge \cdots \ge \mutinf{X;L_m}\\
  \mutinf{Y;L_1}\ge \mutinf{Y;L_2} \ge \cdots \ge \mutinf{Y;L_m}.\label{eq:DPI:Y}
 \end{align}
\end{subequations}
Note that~\eqref{eq:DPI} holds regardless whether the NN is deterministic or stochastic. Nevertheless, several authors have observed violations of the DPI in numerical experiments. The reason for this violation is that the IP does not display mutual information values, but only their estimates. Taking, for example, the binning estimator from~\eqref{eq:plugin}, for every $i$, $Y-X-L_i-Q(L_i)$, but not necessarily $Y-X-Q(L_1)-Q(L_2)-\cdots-Q(L_m)$. Aside from that, also $\mutinf{Y;Q(L_i)}$ needs to be estimated from a finite dataset $\dataset$, so $\hat{p}_{Y,Q(L_i)}$ may not reflect all properties of the joint distribution $P_{Y,Q(L_i)}$ induced by the NN function $f$. Thus, a corresponding chain of inequalities can be assumed to hold at least approximately if the estimated mutual information values are known to be close to the true values. For example, it is reasonable to assume that 
\begin{equation}
\mutest{Y;L_1}\gtrsim \mutest{Y;L_2} \gtrsim \cdots \gtrsim \mutest{Y;L_m}
\end{equation}
if $Q$ has appropriate bin size and the dataset is large enough to ensure that $\mutest{Y;L_i}\approx\mutinf{Y;L_i}$ for every layer. For $\mutest{X;L_i}$, which is known to be a poor estimate of $\mutinf{X;L_i}$ in case the NN is deterministic, there is no reason to believe that a corresponding chain of inequalities holds. 

Indeed, it has been observed that inequalities between $\mutest{X;L}$ of different layers can sometimes be attributed to the layer dimensions. With large bin sizes for the quantizer, narrow layers have a smaller alphabet of $Q(L)$ than wide layers, which effectively bounds $\mutest{X;L}=\entest{Q(L)}$. This is a possible explanation for why $\mutest{X;L}$ decreases with the layer index in a $10-7-5-4-3$ MLP, cf.~\cite{Tishby_BlackBox,Schiemer_RevisitingIP}. In contrast, for a $12-3-2-12-2-2$ MLP the authors of~\cite{Schiemer_RevisitingIP} showed that $\mutest{X;L_3}<\mutest{X;L_4}$ throughout entire training, which is caused by the fact that the third hidden layer has less dimensions than the fourth, thus also less options to map data points to different bins.

Another DPI that holds under~\eqref{eq:DPI:DPI} is $\mutinf{X;L}\ge\mutinf{Y;L}$. For the corresponding estimates, we will have $\mutest{X;L}\gtrsim\mutest{Y;L}$ only if the estimators are comparable. This assumption is inherently problematic as $Y$ is discrete, while $X$ is usually continuously distributed. For example, if $\mutinf{X;L}$ and $\mutinf{Y;L}$ are estimated using KDE with different kernel sizes, then we may have $\mutest{Y;L}>\mutest{X;L}$. Indeed, such a situation occurs in~\cite[Figs.~2~\&~3]{Wickstroem_MatrixEntropies} for $L$ being the softmax layer and a kernel-based estimator (Section~\ref{sec:kernel-based}) and in~\cite[Fig.~3]{Jonsson_CNNs} for layers 13--16 at the end of training for a neural estimator (Section~\ref{sec:neural}). 

\section{Fitting in the Information Plane, Generalization, and Fano's Inequality}\label{sec:fano}
The majority of the literature observes a phase in training NN classifiers during which $\mutest{Y;L}$ increases. This phase is related to empirical risk minimization and is often called \emph{fitting} phase, because the NN learns to fit the labels to the features.

Since in the case of classification, $Y$ usually has a discrete distribution on a finite set of classes, $\mutinf{Y;L_i}$ is finite for every latent representation $L_i$. Furthermore, for every $L_i$ the DPI in~\eqref{eq:DPI:Y} holds. Specifically, if $\hat{Y}$ is the estimate of $Y$ that the NN produces for input $X$, then we have that $\mutinf{X;Y}\ge\mutinf{Y;L_i}\ge\mutinf{Y;\hat{Y}}$ for every latent representation $L_i$. Now suppose that there are $M$ classes that are all equally likely, i.e., we have $p_Y(y)=1/M$ for every class $y$. Then, the classification accuracy $\mathbb{P}(\hat{Y}=Y)$ is bounded from above by the mutual information between $Y$ and $\hat{Y}$~\cite[Th.~1]{Verdu_GeneralizingFano}
\begin{equation}\label{eq:Fano}
 \mutinf{L_i;Y}\ge \mutinf{Y;\hat{Y}}\ge \mathbb{P}(\hat{Y}=Y)\log M + h_2(\mathbb{P}(\hat{Y}=Y))
\end{equation}
where $h_2(\rho)=-\rho\log\rho -(1-\rho)\log(1-\rho)$. Consequently, a high classification accuracy requires large values of $\mutinf{Y;\hat{Y}}$, thus also large values of $\mutinf{L_i;Y}$ for every latent representation.

The converse, however, is not true: If $\mutinf{Y;L}\approx \mutinf{X;Y}$, it need neither be the case that $\hat{Y}=Y$ with high probability nor that $\mutinf{Y;\hat{Y}}$ is large. For example, if $L$ is such that the remaining part of the NN cannot access the information (e.g., by the introduction of an architectural bottleneck between $L$ and subsequent latent representations), then $\mutinf{Y;\hat{Y}}$ may be small despite $L$ being a sufficient statistic for $Y$; cf.~\cite[Section~4.3]{Amjad_LearningRepresentations}. 

Even less clear is the connection between the \emph{estimate} $\mutest{Y;L}$ and classification performance. While large $\mutinf{Y;L}$ is necessary (but not sufficient) for high classification accuracy, the same holds for $\mutest{Y;L}$ only if the estimation scheme and the dataset ensure that $\mutest{Y;L}\approx\mutinf{Y;L}$. Since in principle $\mutest{Y;L}$ can be larger or smaller than $\mutinf{Y;L}$, a large $\mutest{Y;L}$ turns out to be neither necessary nor sufficient for high classification accuracy. Indeed, for $L$ being the softmax output layer and $\mutest{\cdot;\cdot}$ estimated using binning, a three-layer MLP trained on MNIST was shown to have higher $\mutest{Y;L}$ than a six-layer CNN, despite achieving much lower accuracy, cf.~\cite[Tab.~4]{Cheng_ImageClassification}. Furthermore,~\cite[Fig.~4(b)]{Fang_Dissecting} estimated $\mutest{Y;L}\approx 1$ for the last layer of an MLP trained on MNIST using the KDE of~\cite{Kolchinsky_KDE} while claiming an accuracy of 96.93\%. According to~\eqref{eq:Fano}, this means $\mutest{Y;L}$ is a poor estimate of $\mutinf{Y;L}$, either due to bad parameterization or due to insufficient data.

Finally, it is worth mentioning that even though training and validation sets may be drawn from the same joint distribution $P_{X,Y}$, it need not be the case that $\mutest{Y;L}$ is the same when estimated from the training set or the validation set. Indeed, when estimated from the validation set, then overfitting can be seen in the IP as a decrease of $\mutest{Y;L}$ after an initial increase, cf.~\cite[Fig.~3]{Tishby_BlackBox},~\cite[Fig.~5]{Raj_BNNs}, and~\cite{ShwartzZiv_InfiniteWidth}.

\section{Information Plane Analyses of Neural Networks}\label{sec:analyses}
In the following subsections, we critically survey the current literature on IP analyses, categorized according to the type of mutual information estimation. For the sake of readibility, we use $\mutest{\cdot;\cdot}$ for every estimator; critical statements regarding estimation are therefore only valid within the same context (e.g., paragraph, subsection,...). Furthermore, we implicitly assume that the NN under consideration is deterministic unless stated otherwise. 

\subsection{Estimators for Discrete Latent Representations}\label{sec:discrete}
We start with summarizing results for NNs with discrete latent representations. For these, the mutual information terms $\mutinf{X;L}$ and $\mutinf{Y;L}$ are finite and can, at least in principle, be estimated. It is still necessary to ensure that the dataset $\dataset$ is large enough to obtain accurate estimates of $\mutinf{X;L}$ and $\mutinf{Y;L}$, which is particularly challenging if $L$ has a large alphabet, e.g., when the layer defining $L$ consists of many neurons.

For example, the authors of~\cite{Raj_BNNs} investigated binary NNs in which weights are either $+1$ or $-1$ and in which activation functions are step functions, leading to activation values that are either $+1$ or $-1$. The authors applied binning (for the input $X$ and the continuous-valued softmax output) and investigated several techniques for training binary NNs. They performed experiments with a $1024-20-20-20$ MLP on MNIST and a $10-8-6-4-2$ MLP on a binary classification task on the synthetic dataset of dimensionality $4096$ taken from~\cite{Tishby_BlackBox}; we henceforth refer to this dataset as SZT. The authors observed consistently that all layers seem to simultaneously fit (i.e., increase $\mutest{Y;L}$) and compress (i.e., decrease $\mutest{X;L}$), i.e., there is no separate compression phase as claimed in~\cite{Tishby_BlackBox}. Based on experiments with random labels on MNIST, the authors argue that BNNs are not capable of overfitting. While we do not support this conclusion and believe that overfitting can happen in larger binary NNs, the current numerical evidence does not allow drawing conclusions on the interplay between information-theoretic compression and generalization in binary NNs.

The authors of~\cite{Nguyen_IMultiB} also used the SZT dataset to train a stochastic binary NN with four hidden layers of widths $10-8-6-4$. A stochastic binary NNs is characterized by the fact that the conditional distribution of the $\ell$-th neuron in the $i$-th layer is given by
\begin{equation}\label{eq:SNN}
 p_{L_{i,\ell}|L_{i-1}}(1|l) = \sigma(w_{i,\ell} l + b_i)
\end{equation}
where $l$ is the vector of activations of the $(i-1)$-th layer, $w_{i,\ell}$ and $b_i$ are the vector of weights connecting the $\ell$-th neuron in the $i$-th layer with neurons in the $(i-1)$-th layer and the bias term, respectively, and where $\sigma$ is the sigmoid function. Using~\eqref{eq:SNN}, the authors were able to calculate $\mutinf{X;L}$ and $\mutinf{Y;L}$ precisely, i.e., they did not have to resort to estimation. Training these stochastic binary NNs with the aim of minimizing cross entropy does not exhibit compression in the IP. Rather, both $\mutinf{X;L}$ and $\mutinf{Y;L}$ appear to increase throughout training from small initial values, with deeper layers exhibiting a slower increase of $\mutinf{X;L}$ than early layers, cf.~\cite[Fig.~2]{Nguyen_IMultiB}. Interestingly, even when explicitly regularizing NN training using a variant of the IB functional, i.e., even when explicitly encouraging compression in the sense of small $\mutinf{X;L}$, no compression phase in the sense of a reduction of $\mutinf{X;L}$ is visible. 
 
\textbf{Connection to Geometric Compression:} Since both~\cite{Raj_BNNs} and~\cite{Nguyen_IMultiB} provided IPs for NNs with discrete latent representations, the concept of geometric compression in latent space does not apply. Rather, compression visible in the IP can be assumed to be information-theoretic.

\subsection{Binning Estimators}\label{sec:binning}
Binning estimators first apply a quantizer $Q$ to the latent representation $L$ and/or the input $X$ and then compute $\mutest{X;L}$ and $\mutest{Y;L}$ for the binned RVs using the plug-in estimator in~\eqref{eq:plugin}.

The most prominent work in this category is~\cite{Tishby_BlackBox}, in which the authors ran experiments with a $10-7-5-4-3$ MLP with $\tanh$ activation functions trained on SZT via SGD. The quantizer $Q$ was chosen such that it uniformly quantizes the range of each neuron output into 30 bins. The authors observed that, during training, $\mutest{Y;L}$ initially increases fast (fitting phase), while later $\mutest{X;L}$ decreases slowly -- SGD \emph{compressed} information about the features $X$. This led to the claim that SGD runs in two phases, a drift and a diffusion phase, that are purportedly connected with fitting and generalization, respectively. Also, when training with little training data, the network appears to overfit, which is visible in the IP as a late decrease of $\mutest{Y;L}$, cf.~Fig.~\ref{fig:faux} and~\cite[Fig.~3]{Tishby_BlackBox}. 

Subsequently, many claims in~\cite{Tishby_BlackBox} have been challenged. For example, it was shown that compression also occurs when training with full-batch gradient descent~\cite{Saxe_IBTheory}. Thus, SGD is not the (only) reason behind compression in the IP, although it was observed to cause stronger compression than Adam in the softmax layer, cf.~\cite[Fig.~12]{Cheng_ImageClassification}. Similarly, the two-phase nature of NN training has been questioned, as the authors of~\cite{Chelombiev_AdaptiveEstimatorsIB} have observed that, depending on the random initialization of the NN, $\mutest{X;L}$ may first decrease before increasing, first increase before decreasing, only increase, or not change at all. Also the authors of~\cite{Goldfeld_Estimating} observed that $\mutest{X;L}$ may increase and decrease again multiple times throughout training a stochastic $10-7-5-4-3$ MLP.

The interplay between bin size and NN architecture has a great effect on the qualitative picture delivered by the IP, cf.~discussion after~\eqref{eq:plugin:X} and Fig.~\ref{fig:binning}. For example, the authors of~\cite{Schiemer_RevisitingIP} observe that for small bin sizes, many deep layers remain at the point $(\log|\dom{D}|,\mutinf{X;Y})$ in the IP throughout training, indicating that every sample of $\dom{D}$ falls into a different bin. Indeed, for a $1024-20-20-20$ MLP with ReLU activation functions trained on MNIST, the estimates of $\mutinf{X;L}$ obtained via uniform binning with a bin size of $0.5$ converge to $\log|\dataset|$ for every layer, cf.~\cite[Fig.~10]{Saxe_IBTheory}. Similarly, for a CNN with $\tanh$ activation functions trained with Dropout, the authors of~\cite{Goldfeld_Estimating} showed that for all layers $\mutest{X;L}=\entest{Q(L)}\approx \log|\dom{D}|$, despite using only binary quantization.

Also the fact whether compression occurs at all has been questioned by several authors. For example, it has been argued that compression is a consequence of combining the binning estimator with doubly saturating nonlinearities~\cite{Schiemer_RevisitingIP},~\cite[Fig.~2]{Saxe_IBTheory}. For doubly saturating $\tanh$ activation functions, during training less and less bins are used, i.e., the $\tanh$ activation values saturate; this claim is backed by histograms over activation values in~\cite{Saxe_IBTheory,Schiemer_RevisitingIP}. In contrast, the ReLU activation values start at small values initially because of random weight initialization, but then increase to larger values, i.e., the total range of activation values increases, cf.~\cite[Fig.~5]{Schiemer_RevisitingIP}. This leads to a noticeable increase of $\mutest{X;L}$ when $Q$ is a quantizer with a fixed bin size. Consequently, the authors of~\cite{Saxe_IBTheory,Schiemer_RevisitingIP,Goldfeld_Estimating} observe compression in MLPs with $\tanh$ activation functions trained on SZT and MNIST, but not if ReLU activation functions were used. The authors of~\cite{Chelombiev_AdaptiveEstimatorsIB} observed compression for \emph{some} randomly initialized NNs with ReLU activation functions, but not \emph{on average}. NNs with ReLU activation functions display compression in the IP if weight decay is used for regularization~\cite{Chelombiev_AdaptiveEstimatorsIB}, indicating that weight decay has the potential to ``saturate'' the ReLU activation function in the sense of making many neurons inactive. In contrast, regularization encouraging orthonormal weight matrices prevents compression in stochastic NNs with $\tanh$ activation functions~\cite{Goldfeld_Estimating}.

Finally, there has also been evidence that compression may not be linked to generalization~\cite{Saxe_IBTheory,Schiemer_RevisitingIP,Chelombiev_AdaptiveEstimatorsIB,Goldfeld_Estimating}: In~\cite{Saxe_IBTheory}, two deep linear teacher-student networks, one generalizing well and one suffering from overfitting, did not exhibit compression, while compression occurred in a strongly overfitted NN with $\tanh$ activation functions. The authors of~\cite{Schiemer_RevisitingIP} claim that there may be a connection between early stopping and the start of the compression phase, but not between generalization and compression. Indeed, especially for wide hidden layers or convolutional layers it was often observed that $\mutest{X;L}\approx\log|\dataset|$~\cite{Goldfeld_Estimating,Saxe_IBTheory,Li_IBCNN}. In other words, these NNs are almost invertible on the test dataset. Since they are still capable of achieving state-of-the-art results, compression in the IP cannot be necessary for good generalization performance. This is in line with recent results on NNs that are invertible by design~\cite{Jacobsen_iRevnet, Chang_ReversibleNets}. 

In contrast to this, the authors of~\cite{Chelombiev_AdaptiveEstimatorsIB} observe that generalization seems to be correlated with compression in the final softmax layer. The authors of~\cite{Cheng_InfoPlane,Cheng_ImageClassification} arrived at the same conclusion by investigating the IP of a VGG-16 trained on CIFAR-10. This is plausible, as the softmax layer represents the confidence that a sample from $\dataset$ belongs to a certain class. As the NN gains confidence, fewer and fewer quantization patterns become possible, thus reducing $\mutest{X;L}$.

\textbf{Connection to Geometric Compression:} The results of this section are difficult to consolidate: In~\cite{Tishby_BlackBox,Schiemer_RevisitingIP,Goldfeld_Estimating}, the quantizer $Q$ is fixed; for some results in~\cite{Saxe_IBTheory} it is adapted to the training run; and in~\cite{Chelombiev_AdaptiveEstimatorsIB} it is adapted to every layer and every epoch, as bin boundaries are placed at the percentiles of the empirical activation value distribution for each epoch and each layer. In other words, $Q$ is the maximum output entropy quantizer~\cite{Messerschmitt_MOE}. This quantizer is not uniform, as the bins are more narrow in regions where data points accumulate. This makes the results in~\cite{Chelombiev_AdaptiveEstimatorsIB} particularly difficult to interpret. In any case, a reduction of $\mutest{X;L}=\ent{Q(L)}$ is only possible if multiple data points are mapped to the same bin by the NN function $f$. If $Q$ is fixed, one possible cause for many data points being mapped to the same bin is the image of $\dataset$ under $f$ having a small diameter. Another cause is that $f$ maps $\dataset$ to few dense clusters. Both types of geometric compression -- clustering or shrinking -- are thus possible explanations for compression observed in the IP. Simultaneously, clustering is an explanation for fitting, i.e., for an increase of $\mutest{Y;L}$: If $f$ maps samples from each class in $\dataset$ to a different cluster and if $Q$ is such that no two clusters are mapped to the same bin, then $\mutest{Y;L}=\ent{Y}$, which is the maximum achievable value. Thus, geometric clustering according to class membership can explain both fitting and compression phases observed in the IP. Figs.~5a and~8 in~\cite{Goldfeld_Estimating} provide evidence for this explanation by showing that the latent representation $L$ is geometrically clustered in epochs with small $\mutest{X;L}$ and that during training the distances between samples within the same class become significantly smaller than the distances between samples of different classes.

\subsection{Kernel Density Estimation (KDE)}\label{sec:kde}

The authors of~\cite{Saxe_IBTheory} used also KDE to estimate mutual information. Specifically, they relied on the KDE proposed in~\cite{Kolchinsky_KDE}, which assumes the addition of Gaussian noise $\varepsilon$ of variance $\sigma^2$ and provides the following upper and lower bounds:
\begin{multline}\label{eq:kde}
-\frac{1}{|\dataset|} \sum_i \log\frac{1}{|\dataset|} \sum_j \exp\left(-\frac{1}{2} \frac{\Vert f(x_i) - f(x_j)\Vert_2^2}{4\sigma^2}\right) \\
 \le \mutest{X;L} = \mutinf{X;L+\varepsilon} \\
 \le -\frac{1}{|\dataset|} \sum_i \log \frac{1}{|\dataset|} \sum_j \exp\left(-\frac{1}{2} \frac{\Vert f(x_i) - f(x_j)\Vert_2^2}{\sigma^2} \right)
\end{multline}
They admit that ``the addition of noise [in KDE] means that different architectures may no longer be compared in a common currency of mutual information''. And indeed, their results indicate that their upper bound on $\mutest{X;L}$ often lies \emph{below} the value $\mutest{X;L}$ obtained via binning (see~\cite[Figs.~9~\&~10]{Saxe_IBTheory}), for which the DPI implies that $\mutinf{X;L}\ge\mutinf{Q(X);Q(L)}$.

Paralleling our discussion on the effect of the bin size in Section~\ref{sec:binning}, we observe that the variance $\sigma^2$ strongly influences the qualitative picture conveyed by the IP, see Section~\ref{sec:estimation} and Fig.~\ref{fig:KDE}. For example, with $\sigma^2=0.1$, the authors of~\cite{Saxe_IBTheory} showed that compression occurs in a $10-7-5-4-3$ MLP trained on SZT and a $1024-20-20-20$ MLP trained on MNIST only if $\tanh$ activation functions are used. No compression was observed for NNs with ReLU activation functions; for the MNIST setting, $\mutest{X;L}$ eventually converged to $\log|\dataset|$ for all layers. For both ReLU and $\tanh$ activation functions, no compression was visible in~\cite[Fig.~4]{Abrol_MFI} during training a 200-layer MLP on MNIST that was initialized with Gaussian weights. Rather, both $\mutest{X;L}$ and $\mutest{Y;L}$ converged quickly to $\log|\dataset|$ and $\log 10$, respectively, for all but the last layer. In contrast, the authors of~\cite{Chelombiev_AdaptiveEstimatorsIB} observed compression in a later phase of training a $10-7-5-4-3$ MLP with ReLU activation functions on SZT, by adapting, for each epoch, $\sigma^2$ to the maximum activation of the layer. Finally, the authors of~\cite{Fang_Dissecting} used the KDE to estimate the IP for a $1024-256-128$ MLP with ReLU activation functions, a LeNet-5, and a DenseNet, each trained on the MNIST dataset. Their results are inconclusive regarding compression, but all indicate that $\mutest{Y;L}$ increases throughout training, albeit not necessarily monotonically, cf.~\cite[Fig.~4]{Fang_Dissecting}. Interestingly, the authors claim that all three networks achieve test set accuracies exceeding 95\%, indicating $\mutinf{Y;L}\approx\log 10$, cf.~\eqref{eq:Fano}. Yet, even after 50000 epochs, the final layers saturate at around $\mutest{Y;L}=1$, $2$, or $3$ bit, respectively. These facts can only be consolidated by noting that the KDE is a poor estimator of $\mutinf{Y;L}$ in this case, e.g., because the variance $\sigma^2$ of the Gaussian noise $\varepsilon$ was chosen too large.

\textbf{Connection to Geometric Compression:} Looking at the definition of the KDE in~\eqref{eq:kde}, one observes that geometry plays a fundamental role in estimating mutual information: The estimate relies on the pairwise distances between images of data points under the NN function $f$. And indeed, the upper and lower bounds in~\eqref{eq:kde} are tight if the data points are perfectly clustered, i.e., if mapped data points are either very close to each other (within the same cluster) or very far apart (in different clusters)~\cite{Kolchinsky_KDE}. Furthermore, using a fixed variance $\sigma^2$ in~\eqref{eq:kde} applies a fixed measurement scale, and changing the absolute scale of the activation values $L$ relative to this variance leads to compression in the IP; see, e.g.,~\cite[Fig.~1A]{Kolchinsky_KDE}. At the same time, clustering $\dataset$ according to class membership and moving these clusters far apart is a plausible explanation for an increase of $\mutest{Y;L}$. Since in~\cite{Chelombiev_AdaptiveEstimatorsIB} the variance $\sigma^2$ is adapted to the activation values, a decrease of $\mutest{X;L}$ can be explained by clustering, but not by scaling. Thus, clustering appears to be a valid explanation for compression observed in IP analyses based on KDE.

\subsection{Variational and Neural Network-Based Estimators}\label{sec:neural}

Variational estimators of mutual information are obtained by optimizing a parameterized bound on mutual information over a restricted feasible set. For example, a variational lower bound on mutual information is obtained via~\cite[eq.~(2)]{Poole_Variational}
\begin{equation}
 \mutinf{Y;L} = \mathbb{E}\left[\frac{p_{Y|L}(Y|L)}{p_Y(Y)}\right] \ge \mathbb{E}\left[\frac{q_{Y|L}(Y|L)}{p_Y(Y)}\right]
\end{equation}
where the expectation is taken w.r.t.\ the joint distribution of $Y$ and $L$ and where $q_{Y|L}$ is parameterized by, e.g., a NN that is optimized to maximize this lower bound. For example, $q_{Y|L}$ could be a distribution determined by a NN classifier with input $L$, while $q_{X|L}$ in a lower bound on $\mutinf{X;L}$ could be a generative model~\cite{Darlow_ResNet}. Another type of variational lower bound can be obtained from the Donsker-Varadhan representation of mutual information, which is the principle underlying MINE~\cite{Belghazi_MINE}.

MINE was used to estimate the IPs in~\cite{Elad_ValidationIB} and~\cite{Jonsson_CNNs} for a $784-512-512-10$ MLP with ReLU or $\tanh$ activation functions trained on MNIST and a VGG-16 CNN with ReLU activation functions trained on CIFAR-10, respectively. Reference~\cite{Darlow_ResNet} used a NN classifier and a generative PixelCNN++~\cite{Salimans_PixelCNN} as variational distributions for analyzing the IP of a ResNet trained on CINIC-10. While the authors of~\cite{Elad_ValidationIB} added Gaussian noise $\varepsilon$ with variance $\sigma^2=2$ for the purpose of estimation, the other authors appear not to have added noise. The qualitative pictures are entirely different: The authors of~\cite{Jonsson_CNNs} observe compression in deep layers from the first epoch, but no or little compression in early layers; the authors of~\cite{Darlow_ResNet} observe compression at the end of training, but this compression appears to be more pronounced for early layers; and the authors of~\cite{Elad_ValidationIB} do not observe compression at all, unless weight decay is used for regularization, cf.~\cite[Fig.~2 in supplementary material]{Elad_ValidationIB}. All authors observe a fitting phase in the sense that $\mutest{Y;L}$ increases. However, while~\cite[Fig.~3]{Jonsson_CNNs} shows that $\mutest{Y;L}$ increases to $\log 10$ from a comparably large starting value for deeper layers, in~\cite{Elad_ValidationIB} $\mutest{Y;L}$ appears to reach a much smaller value of only 2.2 bit at the end of training.

\textbf{Connection to Geometric Compression:}  These variational estimates are hard to interpret. The images generated using PixelCNN++ from deeper latent representations $L$ are more diverse within the same class than those generated from earlier layers, cf.~\cite[Figs.~4,~5, and~7]{Darlow_ResNet}. Similarly, class-irrelevant features seem to be discarded throughout training. Although the images generated using PixelCNN++ are in good agreement with the IP, it is questionable in how far this admits conclusions regarding the latent representations $L$. Indeed, the authors admit that their estimate $\mutest{X;L}$ is restricted to the ``level of usable information, in as much as it can recover the images''. At the moment, the connection between a reduction of $\mutest{X;L}$ estimated using a generative model and geometric compression remains unclear. Similar considerations hold for the IP in~\cite{Jonsson_CNNs}. Only the authors of~\cite{Elad_ValidationIB} acknowledge that $\mutinf{X;L}=\infty$, in which case MINE is known to yield a poor estimate (\cite[Fig.~1, right]{Belghazi_MINE} or~\cite[Fig.~1]{Elad_ValidationIB}). The authors thus adapt MINE and estimate $\mutinf{X;L+\varepsilon}$ which, with a constant noise variance, is a measure for geometric compression, cf.~Section~\ref{sec:estimation}. And indeed, since small (initial) weights reduce the range of the layer output $L$, this geometric compression explains why $\mutest{X;L}\approx 0$ at the beginning of training and why weight decay leads to compression in the IP.

\subsection{Kernel-Based Estimators}\label{sec:kernel-based}

Reference~\cite{Yu_MatrixEntropies} uses a matrix-based analog of entropy
\begin{subequations}\label{eq:kernel}
 \begin{equation}
 \hat{H}_\alpha(X)=S_\alpha(A)=\frac{1}{1-\alpha} \log\mathrm{tr}(A^\alpha)
\end{equation}
where matrix $A$ is such that $A_{i,j}=\frac{1}{|\dataset|}\frac{G_{i,j}}{\sqrt{G_{i,i}G_{j,j}}}$ and where $G$ is the Gramian obtained by evaluating a Gaussian kernel for all pairs of points in $\dataset$. Shannon entropy is obtained by letting $\alpha\to 1$; the authors have chosen $\alpha=1.01$. Mutual information, e.g., between the input $X$ and a convolutional layer $L=(L_1,\dots,L_c)$ with $c$ filters, is computed by setting
\begin{multline}
 \mutest{X;L} = S_{1.01}(A) + S_{1.01}\left(\frac{A_1\circ\cdots\circ A_c}{\mathrm{tr}(A_1\circ\cdots\circ A_c)}\right)\\
 - S_{1.01}\left(\frac{A\circ A_1\circ\cdots\circ A_c}{\mathrm{tr}(A\circ A_1\circ\cdots\circ A_c)}\right)
\end{multline}
\end{subequations}
where $A_i$ is obtained from the Gramian of $L_i$ and where $\circ$ denotes the Hadamard product. These Hadamard products become numerically problematic for CNNs with many filters, which is why the authors of~\cite{Wickstroem_MatrixEntropies} have proposed the use of tensor kernels in the computation of $A$ instead.

The authors of~\cite{Yu_MatrixEntropies} performed experiments with a LeNet-5 trained on MNIST and FashionMNIST using SGD. Despite using doubly saturating sigmoid activation functions, the authors did not observe any compression phase in the IP. Rather, both $\mutest{X;L}$ and $\mutest{Y;L}$ increase rapidly. It has to be noted, however, that the authors used Silverman's rule-of-thumb to determine the size of the Gaussian kernel required to compute~\eqref{eq:kernel}. This rule-of-thumb has been criticized in~\cite{Tapia_AE} by the fact that the resulting estimates for mutual information are not invariant under linear transforms and that they also depend on the dimensionality of the layer. The authors of~\cite{Tapia_AE} proposed an improved rule for the kernel size and showed that the resulting IPs, computed for autoencoders trained on MNIST, showed better agreement with theoretical considerations.

Using tensor kernels with a learned kernel width, the authors of~\cite{Wickstroem_MatrixEntropies} observe compression only in the softmax layer of a CNN with three convolutional layers and two fully connected layers with widths $400-256$ trained on MNIST while, for the preceding layers, $\mutest{X;L}$ and $\mutest{Y;L}$ stay at high values throughout entire training. In contrast, fitting and subsequent compression were observed in a $1000-20-20-20$ MLP trained on MNIST regardless whether ReLU or $\tanh$ activation functions are used and in deeper layers of a VGG-16 trained on CIFAR-10. Further, the authors of~\cite{Wickstroem_MatrixEntropies} show that at the end of training all layers of the MLP and the CNN have $\mutest{Y;L}\approx \log(10)$, indicating that the NNs may have learned successfully (but see also Section~\ref{sec:fano}). For the VGG-16, however, $\mutest{Y;L}$ is lower when evaluated on the test set than on the training set, which is explained by slight overfitting.

\textbf{Connection to Geometric Compression:} The authors of~\cite{Tapia_AE} showed that there exists a correlation between the mutual information estimated using the approach from~\cite{Yu_MatrixEntropies} and the variance in hidden layers of an autoencoder~\cite[Fig.~8]{Tapia_AE}, suggesting that compression in the IP can be linked to geometric compression, e.g., via simple scaling. Compression observed in the IPs of~\cite{Wickstroem_MatrixEntropies} can also be explained by geometric clustering: If $x_i\approx x_j$ for two data points in the same class, but if $x_i$ and $x_j$ are sufficiently far apart if the class labels are different, then the matrix $A$ is approximately block-diagonal with a rank well approximated by the number of classes (cf. discussion leading to~\cite[eq.~(10)]{Wickstroem_MatrixEntropies}). Therefore, a decrease in $\mutest{X;L}$ can be explained by data points from the same class moving closer together, and data points from different classes moving further apart.

\subsection{Other Estimators}\label{sec:other}

Aside from binning, KDE, kernel-, and NN-based estimation, several other estimation schemes have been proposed and applied to IP analyses of NNs. For example, several authors have used the mutual information estimator proposed in~\cite{Kraskov_Estimator} that is based on $k$-nearest neighbor distances. Indeed, Saxe et al.\ showed that $\mutest{X;L}$ reduces during training a 10-7-5-4-3 MLP on SZT if this MLP has $\tanh$ activation functions, but not if ReLU activation functions are used~\cite[Fig.~12]{Saxe_IBTheory}, thus replicating their conclusions drawn based on binning estimation and KDE. Also the authors of~\cite{Kirsch_ScalableTraining} used the $k$-nearest neighbor estimator from~\cite{Kraskov_Estimator} to estimate $\mutinf{X;L}$, while they used the cross-entropy loss as an estimator of $\mutinf{Y;L}$. The authors experimented with regularization terms encouraging a small second moment of $L$, a small variance of $L$, or a small conditional variance of $L$ given the class variable $Y$. These regularizers thus explicitly encourage $L$ to have a small diameter or to be clustered according to the class variable. The authors performed experiments with a stochastic ResNet18 trained using Adam on CIFAR-10, in which $L=f(X)+\varepsilon$, where $\varepsilon$ is Gaussian noise with identity covariance matrix. The experiments revealed that, for mild regularization, training indeed consisted of a separate compression phase during which $\mutest{X;L}$ decreases (cf.~\cite[Figs.~1 \& Appendix~G]{Kirsch_ScalableTraining}. If regularization is turned off, then compression seems to be absent, indicating that compression is a phenomenon related to the geometric nature of the regularization terms. Furthermore, when evaluated on the test set, overfitting can be seen in the IP by a decrease of $\mutest{Y;L}$ at the end of training, e.g.,~\cite[Fig.~G3]{Kirsch_ScalableTraining}.

While the details of $k$-nearest neighbor estimation depend on the chosen distance metric, the estimates will necessarily reflect geometric phenomena. Indeed, Kirsch et al.\ proposed minimizing the $k$-nearest neighbor estimate of $\diffent{L|Y}+\beta\diffent{L}$ for a ResNet18 on CIFAR-10, providing theoretical evidence that this should lead to a clustered representation $L$, where each cluster corresponds to one class~\cite[Appendix~A2]{Kirsch_EntropyEstimator}. The IP they obtain using $k$-nearest neighbor and NN-based estimates for $\mutinf{X;L}$ and $\mutinf{Y;L}$ indeed show a separate compression phase if $L=f(X)+\varepsilon$, with $\varepsilon$ being Gaussian noise.

Reference~\cite{Gabrie_MI} uses the replica method from statistical physics to estimate the differential entropy (and mutual information) in NNs with wide layers and random, independent and orthogonally-invariant weight matrices. To make the estimate (which is exact in the linear case) finite, they assume that Gaussian noise $\varepsilon$ with variance $\sigma^2=10^{-5}$ was added to $L$, i.e., they estimate $\mutest{X;L}=\mutinf{X;L+\varepsilon}$; the NN is otherwise deterministic. They perform experiments with a $1000-500-250-100$ MLP designed to satisfy the orthogonal invariance of the weight matrices. For NNs with mixed activation functions (linear and ReLU or linear and hardtanh) trained on a synthetic dataset, the authors observe that $\mutest{X;L}$ is always decreasing, potentially after a short initial increase. For NNs with only hardtanh activation functions, the behavior is less consistent and depends on the variance of the initial weights. The source of this compression is hard to determine. Since $\mutest{X;L}$ is an estimate of $\mutinf{X;L+\varepsilon}$, a certain geometric influence cannot be ruled out. In any case, since an eventual compression is not synchronized with generalization, Gabri\'e et al.\ conclude that compression and generalization may not be linked and that within their examined setting ``a simple information theory of deep learning [may remain] out-of-reach''.

In~\cite{ShwartzZiv_InfiniteWidth}, the authors consider infinite ensembles of infinitely wide deterministic NNs initialized with Gaussian weights, for which it is shown that minimizing the square loss makes the distribution $P_{L|X}$ Gaussian for every epoch. Thus, the ensemble is stochastic with finite mutual information values and they estimate $\mutinf{X;L}$ and $\mutinf{Y;L}$ via multi-sample and variational bounds. Their experiments with an ensemble of $1000-1000-1000$ MLPs trained on a Gaussian regression dataset do not show compression in the IP, but indicate that overfitting is represented by a decrease of $\mutest{Y;L}$ combined with an increase of $\mutest{X;L}$. Overfitting seems to be more severe in NNs with $\mathrm{erf}$ activation functions than in those with ReLU activation functions, and it seems to be more severe in NNs initialized with small weight variances. Similar conclusions regarding overfitting and the impact of the weight variance were obtained for training a two-layer CNN to determine the parity of MNIST digits, albeit on the basis of different qualitative behavior in the IPs, cf.~\cite[Figs.~2 and~3]{ShwartzZiv_InfiniteWidth}.

The authors of~\cite{Balda_ITView,Balda_Trajectory} considered $L=\hat{Y}$ to be the output of the NN, i.e., $L$ is the index of the largest activation in the final layer. Since $L$ has finite alphabet, $\mutinf{X;L}=\ent{L}$ and $\mutinf{Y;L}$ are finite and can be estimated reliably using the plug-in estimator~\eqref{eq:plugin}. The authors observed that NN training consists of two phases that are entirely different from those in~\cite{Tishby_BlackBox}. For all considered architectures, training sets, and activation functions, the authors first observed a strong increase of $\mutest{X;L}=\entest{L}$ followed by a decrease of $\entest{L|Y}$, i.e., $\mutest{Y;L}$ increases more strongly than $\entest{L}$ in the second phase. Compression does not happen at all: The authors prove that $\entest{L}$ is bounded from below by an increasing function of training time, cf.~\cite[Prop.~3]{Balda_Trajectory}. In other words, the output of the NN learns about the input throughout training, but starts learning towards the task only if the output contains sufficient information $\entest{L}$; the onset of task-specific learning occurs at larger values of $\entest{L}$ if label noise is added.

The authors of~\cite{Goldfeld_Estimating} used the sample propagation estimator from~\cite{Goldfeld_Estimator} to estimate the (finite) mutual information $\mutinf{X;L}$ in stochastic NNs, where independent and identically distributed Gaussian noise is added to every neuron output. The estimator was shown to yield qualitatively similar results as a binning estimator of the entropy of $L$, i.e., compression in the IP can be explained by geometric compression of the latent representation $L$.

Finally, the authors of~\cite{Noshad_EDGE} proposed an estimator based on collisions of a locality sensitive hash function. Specifically, they use the plug-in estimators in~\eqref{eq:plugin}, where the quantizer $Q$ is a composition of uniform binning and a random non-injective hash function with a uniform density on its range. In their experiments with a $1024-20-20-20$ MLP trained on MNIST, they observed compression in the hidden layers for both $\tanh$ and ReLU activation functions. Compression was also observed in a $200-100-60-30$ MLP and a CNN with ReLU activation functions. For the CNN, the authors did not observe a significant fitting phase, i.e., it appears as if $\mutest{Y;L}$ is large already at the beginning of training or after very few epochs. Since binning plays a major role also for this estimator -- points in latent space that are close to each other are mapped to the same hash value -- geometric compression is again a possible explanation for the observed reduction of $\mutest{X;L}$.

\section{Discussion: Geometric Compression and the Utility of the Information Plane}
\label{sec:summary}

In this survey, we have discussed the existing literature on IP analyses and tried to interpret the main results from the perspective of mutual information estimation. Our main conclusions from this perspective are the following:

\begin{enumerate}
 \item \label{item:1} In deterministic NNs, the mutual information $\mutinf{X;L}$ between the input $X$ and the latent representation $L$ is infinite. Thus, any finite value of $\mutest{X;L}$ depends strongly on the method of estimation. Since there is no consensus yet on how the mutual information values in the IP shall be estimated, it follows that IPs produced with different estimators cannot be compared (Section~\ref{sec:estimation}).
 
 \item As an example for the influence of the estimation method, it can be seen that the DPI is not always satisfied for the estimates $\mutest{X;L}$ and $\mutest{Y;L}$. The literature either acknowledges that explicitly, or this conclusion can be drawn from the presented evidence. Taking into account that $\mutest{\cdot;\cdot}$ rarely coincides with $\mutinf{\cdot;\cdot}$, this discrepancy is not surprising (Section~\ref{sec:DPI}).
 
 \item Since the literature reaches consensus about certain qualitative properties of the IP, it follows that different estimation methods must share certain common properties. Based on both theoretical considerations (Section~\ref{sec:estimation}), reference to the literature~\cite{Goldfeld_Review,Goldfeld_Estimating,Schiemer_RevisitingIP}, and the interpretation of existing IP analyses, we conclude that this common property is related to \emph{geometric} phenomena.

 \item With reference to item~\ref{item:1}, compression in the IP cannot be information-theoretic for deterministic NNs. Rather, for most estimators it is compatible with geometric compression in the sense that the NN function $f$ maps $\dataset$ to clusters in latent space, or that the diameter of $f(\dataset)$ is small. Whether clustering or shrinking is the cause behind small $\mutest{X;L}$ depends on the details of estimation: Doubly saturating activation functions encourage clustering because activation values saturate, while weight decay encourages small weights, effectively restricting the diameter of $f(\dataset)$. For stochastic NNs in which noise is added to neuron outputs, $\mutest{X;L}$ may measure both information-theoretic and geometric compression~\cite{Goldfeld_Estimating}. This conclusion is supported by considering results of training stochastic NNs with approximations of $\mutinf{X;L}$ as a regularization term. In such NNs, the latent representations indeed appear clustered, cf.~\cite[Fig.~2]{Alemi_DVIB} and~\cite[Fig.~1]{Kolchinsky_NLIB}.
 
 \item The majority of the literature observes a fitting phase in NN training during which $\mutest{Y;L}$ increases. While NNs with good generalization performance must have large $\mutinf{Y;L}$ as dictated by Fano's inequality and the DPI, neither does large $\mutinf{Y;L}$ guarantee good generalization performance, nor does good generalization performance guarantee large $\mutest{Y;L}$ (Section~\ref{sec:fano}). The observed fitting phase may thus be either due to $\mutest{Y;L}\approx\mutinf{Y;L}<\infty$ or due to the fact that the employed estimators capture the same geometric phenomenon (e.g., geometric separation of classes in latent space).
 
 \item Several authors have observed that overfitting appears in the IP as an increase of $\mutest{Y;L}$ followed by a decrease of $\mutest{Y;L}$, when these quantities are estimated from a test or validation set.
 
 \item\label{item:nocompression} A compression phase in NN training during which $\mutest{X;L}$ decreases is not as common as initial claims would suggest, cf.~Table~\ref{tab:literature}.
 
 \item\label{item:CNNs} For example, the convolutional layers of CNNs appear to behave as invertible functions $f$ at all training epochs: $\mutest{Y;L}$ is large and $\mutest{X;L}$ is close to $\log|\dataset|$ throughout entire training. Similar observations were made on other high-dimensional latent representations $L$, e.g., where $L$ is the output of a wide hidden layer in a MLP. Invertibility is thus not in conflict with good generalization performance, cf.~\cite{Jacobsen_iRevnet,Chang_ReversibleNets}.
 
 \item As a consequence from items~\ref{item:nocompression} and~\ref{item:CNNs}, for deterministic NNs, it is neither sufficient nor necessary for good generalization performance that $L$ is a \emph{minimum} sufficient statistic for $Y$, with minimality measured by $\mutinf{X;L}$. 
 
 \item Good generalization performance may, however, be correlated with small $\mutest{X;L}$ -- according to Table~\ref{tab:literature}, this is rarely the case. If there is a correlation, the reason for good generalization performance shall be sought through an interpretation of the estimator, rather than in the claim that $L$ is minimum sufficient statistic. An in-depth understanding of the meaning of $\mutest{X;L}$ in the respective scenario may lead to novel regularization approaches for NN training,~cf.~\cite{Elad_ValidationIB,Kolchinsky_NLIB,Alemi_DVIB}.
 
 \item While compression in the IP can be reasonably explained by geometric compression, geometric compression is not necessary for good generalization performance. For example, the setting visualized in~\cite[Fig.~5b]{Goldfeld_Estimating} achieves top classification performance without geometric clustering. Whether class-specific clustering is sufficient for good classification performance is unclear, but seems plausible.

\end{enumerate}

The summary of these conclusions is that the IP does not depict what was initially expected. More specifically, compression in the IP does not necessarily represent learning a minimum sufficient statistic, nor does the concept of a minimum sufficient statistic seem to be necessary for good generalization performance. Since similarly geometric compression (which can be assumed to appear as compression in the IP) seems to be neither necessary nor sufficient for good generalization performance, the purported connection between compression in the IP and generalization performance seems highly questionable given the current state of knowledge, cf.~Table~\ref{tab:literature}.

Nevertheless, we believe that the IP has the potential to ``open the black box of deep learning'' in certain aspects. For example, fitting and overfitting appear to have counterparts in the IP. Taking the properties of mutual information estimation into account, IP analyses are a valid way to investigate geometric phenomena during NN training, such as geometric compression or class separation; a recent work exploiting this connection is~\cite{Basirat_IPlane}. Being able to summarize these geometric properties of an entire layer with just two numbers per epoch, the IP thus yields insights into the effects of regularization and learning schemes that are not available from test set accuracies or learning curves. Finally, if the NNs under investigation are such that $\mutinf{X;L}$ is finite, as in stochastic NNs or binary NNs, and if the estimators are parameterized appropriately and the dataset $\dataset$ is sufficiently large, then $\mutest{\cdot;L}$ can be an adequate estimate of the finite mutual information $\mutinf{\cdot;L}$. In such a case, the IP depicts an inherently information-theoretic picture. While currently only few references fall into this category, e.g.,~\cite{Goldfeld_Estimating,Raj_BNNs,Nguyen_IMultiB}, we are convinced that putting more effort into investigating the information-theoretic learning behavior is capable of shedding more light on the inner workings of NNs.

\bibliographystyle{IEEEtran}
\bibliography{IEEEabrv,../../LiteratureSurvey/network_pruning}

\begin{IEEEbiography}[{\includegraphics[width=1in,height=1.25in,clip,keepaspectratio]{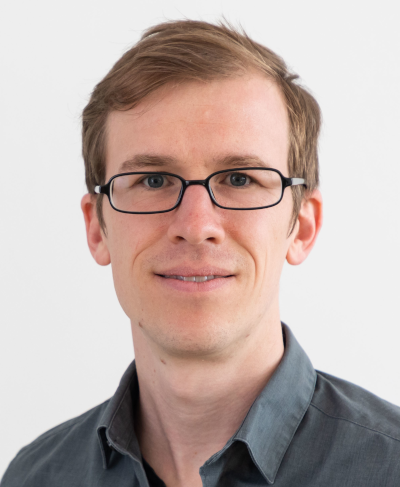}}]{Bernhard C. Geiger}
(S'07, M'14, SM'19) received the Dipl.-Ing. degree in electrical engineering (with distinction) and the Dr. techn. degree in electrical and information engineering (with distinction) from Graz University of Technology, Austria, in 2009 and 2014, respectively.

In 2009 he joined the Signal Processing and Speech Communication Laboratory, Graz University of Technology, as a Project Assistant and took a position as a Research and Teaching Associate at the same lab in 2010. He was a Senior Scientist and Erwin Schr\"odinger Fellow at the Institute for Communications Engineering, Technical University of Munich, Germany from 2014 to 2017 and a postdoctoral researcher at the Signal Processing and Speech Communication Laboratory, Graz University of Technology, Austria from 2017 to 2018. He is currently a Senior Researcher at Know-Center GmbH, Graz, Austria, where he also leads the Machine Learning Group within the Knowledge Discovery Area. His research interests cover information theory for machine learning, theory-assisted machine learning, and information-theoretic model reduction for Markov chains and hidden Markov models.
\end{IEEEbiography}

\end{document}

%% file: table.tex
\begin{table*}[t]
\caption{Overview of the surveyed literature. The last two columns indicate whether compression (comp) was observed ($\surd$), was not observed ($\times$), or whether the picture was inconsistent ($\sim$); and whether the authors made supporting ($\surd$), negating ($\times$), or both types ($\sim$) of claims, or no claim at all (empty) for a causal link between compression and generalization (gen). It can be seen that compression is less common than expected, and that its connection with generalization has been questioned. We used the following abbreviations: batch gradient descent (BGD), binarized NN (BNN), binning (bin), DenseNet (DN), FashionMNIST (F-MNIST), $k$-nearest neighbor ($k$-nn), LeNet (LN), sigmoid (sig.), stochastic gradient descent (SGD), stochastic binary NN (SNN), synthetic dataset (syn.), the dataset from~\cite{Tishby_BlackBox} (SZT), training method (Train.), variational (var.). The term ``\textsc{all}'' in the column for the training method indicates that SGD, BGD, and Adam were used for training in different experiments.
}
\label{tab:literature}
\begin{center}
\begin{scriptsize}
\begin{sc}
\begin{tabular}{lccccccc}
\toprule
    Reference
    & Architecture            
    & Activation                  
    & Train. 
    & Dataset
    & Estimator
    & comp
    & gen\\
\midrule

Abrol \& Tanner~\cite{Abrol_MFI}
& MLP
& $\tanh$, ReLU
& ?
& MNIST
& KDE
& $\times$ 
& \\

Balda et al.~\cite{Balda_ITView,Balda_Trajectory}
& MLP, LN, DN
& $\tanh$, sig., ReLU
& SGD
& MNIST, CIFAR, syn.
& decision rule
& $\times$ 
& \\

Chelombiev et al.~\cite{Chelombiev_AdaptiveEstimatorsIB}  
&  MLP
& $\tanh$, ReLU 
& Adam
& SZT 
& bin, KDE
& $\sim$ 
& $\sim$ \\

Cheng et al.~\cite{Cheng_ImageClassification}  
& MLP, CNN
& softmax 
& all
& MNIST, CIFAR-10
&  bin
& $\sim$ 
& $\surd$ \\

Darlow \& Storkey~\cite{Darlow_ResNet}  
& ResNet
& leaky ReLU 
& SGD
& CINIC-10
& var.
& $\surd$ \\

Elad et al.~\cite{Elad_ValidationIB}
& MLP
& $\tanh$, ReLU 
& Adam
& MNIST
& MINE~\cite{Belghazi_MINE}
& $\sim$ \\

Fang et al.~\cite{Fang_Dissecting}  
&  MLP, LN, DN
& ReLU 
& Adam
& MNIST 
& KDE
& $\sim$ \\

Gabri\'e et al.~\cite{Gabrie_MI}
& MLP
& hardtan, ReLU  
& SGD
& syn.
& replica
& $\sim$ 
& $\times$ \\

Goldfeld et al.~\cite{Goldfeld_Estimating}  
&  MLP, CNN   
&$\tanh$, ReLU 
& SGD
& SZT, MNIST, syn. 
&  bin,~\cite{Goldfeld_Estimator}
& $\sim$
& $\times$ \\

J\'onsson et al.~\cite{Jonsson_CNNs}
& VGG16
& ReLU 
& ?
& CIFAR-10
& MINE~\cite{Belghazi_MINE}
& $\surd$ \\

Kirsch et al.~\cite{Kirsch_ScalableTraining}
& ResNet18
& ? 
& Adam
& CIFAR-10
& $k$-nn~\cite{Kraskov_Estimator}, var.
& $\surd$ \\

Nguyen \& Choi~\cite{Nguyen_IMultiB}
& SNN
& $-$
& SGD
& SZT
& exact
& $\times$ \\

Noshad et al.~\cite{Noshad_EDGE}
& MLP, CNN
& $\tanh$, ReLU  
& Adam
& MNIST
& bin + hash
& $\surd$ \\

Raj et al.~\cite{Raj_BNNs}
& BNN
& $-$
& Adam
& MNIST, SZT
& bin
& $\surd$
& \\

Shwartz-Zin \& Tishby~\cite{Tishby_BlackBox}      
&  MLP          
&$\tanh$                
& SGD
& SZT 
&  bin  
& $\surd$ 
& $\surd$ \\

Saxe et al.~\cite{Saxe_IBTheory}        
& MLP           
&$\tanh$, ReLU, lin.   
& SGD, BGD
& SZT, MNIST, syn. 
&  bin, KDE  
& $\sim$ 
& $\times$ \\

Schiemer \& Ye~\cite{Schiemer_RevisitingIP}  
&  MLP
&$\tanh$, ReLU 
& SGD
& SZT, MNIST 
&  bin
& $\sim$ 
& $\times$ \\

Schwartz-Ziv \& Alemi~\cite{ShwartzZiv_InfiniteWidth}
& MLP, CNN
& erf, ReLU
& ?
& MNIST, syn.
& var., sample
& $\times$
& $\times$ \\

Wickstr\o{}m et al.~\cite{Wickstroem_MatrixEntropies}
& MLP, CNN, VGG16
& $\tanh$, ReLU  
& SGD
& MNIST, CIFAR-10
& kernel
& $\sim$
& $\times$ \\

Yu et al.~\cite{Yu_MatrixEntropies}
& LN
& sig. 
& SGD
& MNIST, F-MNIST
& kernel
& $\times$

\end{tabular}
\end{sc}
\end{scriptsize}
\end{center}
\vskip -0.1in
\end{table*}